%% file: main2.tex
\documentclass[11pt]{article}
\usepackage{enumerate}
\usepackage[OT1]{fontenc}
\usepackage{amsmath,amssymb}
\usepackage{natbib}
\usepackage[usenames]{color}

\usepackage{smile}
\usepackage{multirow}
\usepackage[breaklinks,colorlinks,
            linkcolor=red,
            anchorcolor=blue,
            citecolor=blue
            ]{hyperref}

\def\given{\,|\,}
\def\biggiven{\,\big{|}\,}
\def\tr{\mathop{\text{tr}}\kern.2ex}

\RequirePackage{amsmath}
\RequirePackage{amssymb}
\RequirePackage{amsthm}
\RequirePackage{bm}
\RequirePackage{url}
\usepackage{multirow}
\usepackage{natbib}
\usepackage{graphicx}
\usepackage{subfigure}
\usepackage{makecell}
\usepackage{booktabs}
\usepackage{array}
\usepackage{url}
\usepackage{algorithm}
\usepackage{algorithmic}
\usepackage{dsfont}
\usepackage{mathtools}
\usepackage{algorithmic}
\usepackage{tikz}

\newcommand{\iid}{{\mathrm{i.i.d.}\!}}

\newcommand{\rd}{{\mathrm{d}}}
\newcommand{\rE}{{\text{{\tiny E}}}}

\def\sP{\mathscr{P}}
\def\sB{\mathscr{B}}
\def\lip{{\rm Lip}}
\def\softmax{\mathop{\mathrm{softmax}}}

\long\def\comment#1{}

\DeclarePairedDelimiterX{\norm}[1]{\|}{\|}{#1}

\def\msbe{\text{MSBE}}
\def\mspbe{\text{MSPBE}}
\def\Div{\mathop{\mathrm{div}}}

\def\law{\mathop{\mathrm{law}}}
\def\sigmoid{\mathop{\mathrm{sigmoid}}}

\let\hat\widehat
\let\tilde\widetilde

\def\given{{\,|\,}}
\def\biggiven{{\,\big|\,}}
\def\Biggiven{{\,\Big|\,}}
\def\bigggiven{{\,\bigg|\,}}
\def\Bigggiven{{\,\Bigg|\,}}

\DeclarePairedDelimiterX{\inp}[2]{\langle}{\rangle}{#1, #2}

\long\def\comment#1{}

\def\tr{\mathop{\text{Tr}}}

\def\cS{{\mathcal{S}}}

\def\cX{{\mathcal{X}}}
\def\cD{{\mathcal{D}}}

\def\cL{{\mathcal{L}}}

\def\cT{{\mathcal{T}}}
\def\cB{{\mathcal{B}}}
\def\tr{{\text{Tr}}}

\providecommand{\norm}[1]{\vvvert#1\vvvert}

\newcommand{\bel}{\begin{eqnarray}\label}
\newcommand{\eel}{\end{eqnarray}}
\newcommand{\bes}{\begin{eqnarray*}}
\newcommand{\ees}{\end{eqnarray*}}

\usepackage{mathrsfs}

\usepackage{fullpage}

\makeatletter
\renewenvironment{abstract}
{%
	{\list{}{\addtolength{\leftmargin}{-2em} 
			\listparindent 0em%
			\itemindent    \listparindent%
			\rightmargin   \leftmargin%
			\parsep        \z@ \@plus\p@}%
		\item\relax}%
	{\endlist}%
	\oldabstract}
{\oldendabstract}
\makeatother

\usepackage{hyperref}
\usepackage[english]{babel}
\usepackage{breakcites}
\usepackage{microtype}
\def\##1\#{\begin{align}#1\end{align}}
\def\$#1\${\begin{align*}#1\end{align*}}
\usepackage{enumitem}

\usepackage{setspace} 

\begin{document}

\title{\Huge Can Temporal-Difference and Q-Learning\\ Learn Representation? A Mean-Field Theory}

\author{
	\normalsize Yufeng Zhang\thanks{Northwestern University; \texttt{yufengzhang2023@u.northwestern.edu}}
	\qquad
	\normalsize Qi Cai\thanks{Northwestern University; \texttt{qicai2022@u.northwestern.edu}}
	\qquad
	\normalsize Zhuoran Yang\thanks{Princeton University; \texttt{zy6@princeton.edu}} 
	\qquad
	\normalsize Yongxin Chen\thanks{Georgia Institute of Technology; \texttt{yongchen@gatech.edu}}
	\qquad
	\normalsize Zhaoran Wang\thanks{Northwestern University; \texttt{zhaoranwang@gmail.com}}
}
\date{}
\maketitle

\input{intro}

\input{background}

\input{algo}

\input{result}

\input{proof}

\input{extension}


\section{Acknowledgement}
Zhaoran Wang acknowledges National Science Foundation (Awards 2048075, 2008827, 2015568, 1934931), Simons Institute (Theory of Reinforcement Learning), Amazon, J.P. Morgan, and Two Sigma for their supports.

\bibliographystyle{ims}
\bibliography{graphbib}

\input{appendix}

\end{document}

%% file: intro.tex

\begin{abstract}

Temporal-difference and Q-learning play a key role in deep reinforcement learning, where they are empowered by expressive nonlinear function approximators such as neural networks. At the core of their empirical successes is the learned feature representation, which embeds rich observations, e.g., images and texts, into the latent space that encodes semantic structures. Meanwhile, the evolution of such a feature representation is crucial to the convergence of temporal-difference and Q-learning. 

\vskip8pt
In particular, temporal-difference learning converges when the function approximator is linear in a feature representation, which is fixed throughout learning, and possibly diverges otherwise. We aim to answer the following questions: {\it When the function approximator is a neural network, how does the associated feature representation evolve? If it converges, does it converge to the optimal one?}~~

\vskip8pt
We prove that, utilizing an overparameterized two-layer neural network, temporal-difference and~Q-learning globally minimize the mean-squared projected Bellman error at a sublinear rate. Moreover, the associated feature representation converges to the optimal one, generalizing the previous analysis of \cite{cai2019neural} in the neural tangent kernel regime, where the associated feature representation stabilizes at the initial one. The key to our analysis is a mean-field perspective, which connects the evolution of a finite-dimensional parameter to its limiting counterpart over an infinite-dimensional Wasserstein space. Our analysis generalizes to soft Q-learning, which is further connected to policy gradient.

\end{abstract}

\newpage

\section{Introduction}
Deep reinforcement learning achieves phenomenal empirical successes, especially in challenging applications where an agent acts upon rich observations, e.g., images and texts. Examples include video gaming \citep{mnih2015human}, visuomotor manipulation \citep{levine2016end}, and language generation \citep{he2015deep}. Such empirical successes are empowered by expressive nonlinear function approximators such as neural networks, which are used to parameterize both policies (actors) and value functions (critics) \citep{konda2000actor}. In particular, the neural network learned from interacting with the environment induces a data-dependent feature representation, which embeds rich observations into a latent space encoding semantic structures \citep{hinton1986learning, bengio2012deep, yosinski2014transferable, lecun2015deep}. In contrast, classical reinforcement learning mostly relies on a handcrafted feature representation that is fixed throughout learning \citep{sutton2018reinforcement}. 

In this paper, we study temporal-difference (TD) \citep{sutton1988learning} and Q-learning \citep{watkins1992q}, two of the most prominent algorithms in deep reinforcement learning, which are further connected to policy gradient \citep{williams1992simple} through its equivalence to soft Q-learning \citep{o2016combining, schulman2017equivalence, nachum2017bridging, haarnoja2017reinforcement}. In particular, we aim to characterize how an overparameterized two-layer neural network and its induced feature representation evolve in TD and Q-learning, especially their rate of convergence and global optimality. A fundamental obstacle, however, is that such an evolving feature representation possibly leads to the divergence of TD and Q-learning. For example, TD converges when the value function approximator is linear in a feature representation, which is fixed throughout learning, and possibly diverges otherwise \citep{baird1995residual,boyan1995generalization,tsitsiklis1997analysis}. 

To address such an issue of divergence, nonlinear gradient TD \citep{bhatnagar2009convergent} explicitly linearizes the value function approximator locally at each iteration, that is, using its gradient with respect to the parameter as an evolving feature representation. Although nonlinear gradient TD converges, it is unclear whether the attained solution is globally optimal. On the other hand, when the value function approximator in TD is an overparameterized multi-layer neural network, which is required to be properly scaled, such a feature representation stabilizes at the initial one \citep{cai2019neural}, making the explicit local linearization in nonlinear gradient TD unnecessary. Moreover, the implicit local linearization enabled by overparameterization allows TD (and Q-learning) to converge to the globally optimal solution. However, such a required scaling, also known as the neural tangent kernel (NTK) regime \citep{jacot2018neural}, effectively constrains the evolution of the induced feature presentation to an infinitesimal neighborhood of the initial one, which is not data-dependent.

\vskip4pt

\noindent{\bf Contribution.} Going beyond the NTK regime, we prove that, when the value function approximator is an overparameterized two-layer neural network, TD and Q-learning globally minimize the mean-squared projected Bellman error (MSPBE) at a sublinear rate. Moreover, in contrast to the NTK regime, the induced feature representation is able to deviate from the initial one and subsequently evolve into the globally optimal one, which corresponds to the global minimizer of the MSPBE. We further extend our analysis to soft Q-learning, which is connected to policy gradient. 

The key to our analysis is a mean-field perspective, which allows us to associate the evolution of a finite-dimensional parameter with its limiting counterpart over an infinite-dimensional Wasserstein space \citep{villani2003topics,villani2008optimal,ambrosio2008gradient,ambrosio2013user}. Specifically, by exploiting the permutation invariance of the parameter, we associate the neural network and its induced feature representation with an empirical distribution, which, at the infinite-width limit, further corresponds to a population distribution. The evolution of such a population distribution is characterized by a partial differential equation (PDE) known as the continuity equation. In particular, we develop a generalized notion of one-point monotonicity \citep{harker1990finite}, which is tailored to the Wasserstein space, especially the first variation formula therein \citep{ambrosio2008gradient}, to characterize the evolution of such a PDE solution, which, by a discretization argument, further quantifies the evolution of the induced feature representation.  


\vskip4pt

\noindent{\bf Related Work.} When the value function approximator is linear, the convergence of TD is extensively studied in both continuous-time \citep{jaakkola1994convergence,tsitsiklis1997analysis,borkar2000ode,kushner2003stochastic,borkar2009stochastic} and discrete-time \citep{bhandari2018finite,lakshminarayanan2018linear,dalal2018finite,srikant2019finite} settings. See \cite{dann2014policy} for a detailed survey. Also, when the value function approximator is linear, \cite{melo2008analysis, zou2019finite, chen2019performance} study the convergence of Q-learning. When the value function approximator is nonlinear, TD possibly diverges \citep{baird1995residual,boyan1995generalization,tsitsiklis1997analysis}. \cite{bhatnagar2009convergent} propose nonlinear gradient TD, which converges but only to a locally optimal solution. See \cite{geist2013algorithmic, bertsekas2019feature} for a detailed survey. When the value function approximator is an overparameterized multi-layer neural network, \cite{cai2019neural} prove that TD converges to the globally optimal solution in the NTK regime. See also the independent work of \cite{brandfonbrener2019geometric, brandfonbrener2019expected, agazzi2019temporal, sirignano2019asymptotics}, where the state space is required to be finite. In contrast to the previous analysis in the NTK regime, our analysis allows TD to attain a data-dependent feature representation that is globally optimal. 

Meanwhile, our analysis is related to the recent breakthrough in the mean-field analysis of stochastic gradient descent (SGD) for the supervised learning of an overparameterized two-layer neural network \citep{chizat2018global,mei2018mean,mei2019mean, javanmard2019analysis,wei2019regularization,fang2019over,fang2019convex,chen2020mean}. See also the previous analysis in the NTK regime \citep{daniely2017sgd, chizat2018note, jacot2018neural, li2018learning, allen2018learning, allen2018convergence, du2018gradientb, du2018gradient, zou2018stochastic, arora2019exact, arora2019fine, lee2019wide, cao2019bounds, chen2019much, zou2019improved, ji2019polylogarithmic, bai2019beyond}. Specifically, the previous mean-field analysis casts SGD as the Wasserstein gradient flow of an energy functional, which corresponds to the objective function in supervised learning. In contrast, TD follows the stochastic semigradient of the MSPBE \citep{sutton2018reinforcement}, which is biased. As a result, there does not exist an energy functional for casting TD as its Wasserstein gradient flow. Instead, our analysis combines a generalized notion of one-point monotonicity \citep{harker1990finite} and the first variation formula in the Wasserstein space \citep{ambrosio2008gradient}, which is of independent interest. 

\vskip4pt

\noindent{\bf Notations.} We denote by $\sB(\cX)$ the Borel $\sigma$-algebra over the space $\cX$. Let $\sP(\cX)$ be the set of Borel probability measures over the measurable space $(\cX, \sB(\cX))$. We denote by $[N] = \{1, 2, \ldots, N\}$ for any $N\in \NN_+$. Also, we denote by $\cB^n(x; r) =\{y\in \RR^n\given \norm{y-x} \le r \}$ the closed ball in $\RR^n$. Given a curve $\rho:\RR\rightarrow \cX$, we denote by $\rho'_s = \partial_t \rho_t\given_{t=s}$ its derivative with respect to the time. For a function $f: \cX \rightarrow \RR$, we denote by $\lip(f) = \sup_{x, y\in \cX, x\neq y} |f(x) - f(y)| / \norm{x- y}$ its Lipschitz constant. For an operator $F: \cX \rightarrow \cX$ and a measure $\mu \in \sP(\cX)$, we denote by $F_\sharp \mu = \mu \circ F^{-1}$ the push forward of $\mu$ through $F$. We denote by $D_{\rm KL}$ and $D_{\chi^2}$ the Kullback-Leibler (KL) divergence and the $\chi^2$ divergence, respectively.

%% file: background.tex
\section{Background}
\subsection{Policy Evaluation}
We consider a Markov decision process $(\cS, \cA, P, R, \gamma, \cD_0)$, where $\cS\subseteq \RR^{d_1}$ is the state space, $\cA \subseteq \RR^{d_2}$ is the action space, $P: \cS\times \cA\rightarrow \sP(\cS)$ is the transition kernel, $R: \cS\times \cA\rightarrow \sP(\RR)$ is the reward distribution, $\gamma \in (0, 1)$ is the discount factor, and $\cD_0 \in \sP(\cS)$ is the initial state distribution. An agent following a policy $\pi: \cS\rightarrow \sP(\cA)$ interacts with the environment in the following manner. At a state $s_t$, the agent takes an action $a_t$ according to $\pi(\cdot \given s_t)$ and receives from the environment a random reward $r_t$ following $R(\cdot \given s_t, a_t)$. Then, the environment transits into the next state $s_{t+1}$ according to $P(\cdot \given s_t, a_t)$. We measure the performance of a policy $\pi$ via the expected cumulative reward $J(\pi)$, which is defined as follows,
\begin{align}
\label{eq:def-j}
J(\pi) = \EE\Bigl[ \sum_{t=0}^\infty \gamma^t \cdot r_t \Biggiven s_0 \sim \cD_0, a_t \sim \pi(\cdot\given s_t), r_t\sim R(\cdot\given s_t, a_t), s_{t+1} \sim P(\cdot\given s_t, a_t) \Bigr].
\end{align}
In policy evaluation, we are interested in the state-action value function (Q-function) $Q^\pi: \cS\times \cA \rightarrow \RR$, which is defined as follows,
\begin{align*}
Q^\pi(s, a) &= \EE\Bigl[ \sum_{t=0}^\infty \gamma^t \cdot r_t \Biggiven s_0 =s, a_0 = a,  a_t \sim \pi(\cdot\given s_t), r_t\sim R(\cdot\given s_t, a_t), s_{t+1} \sim P(\cdot\given s_t, a_t) \Bigr].
\end{align*}
We learn the Q-function by minimizing the mean-squared Bellman error  (MSBE), which is defined as follows,
\$\msbe(Q) = \frac{1}{2} \cdot \EE_{(s, a) \sim \cD} \Bigl[\bigl( Q(s, a) - \cT^\pi Q(s, a) \bigr)^2\Bigr].\$
Here $\cD \in \sP(\cS\times \cA)$ is the stationary distribution induced by the policy $\pi$ of interest and $\cT^\pi$ is the corresponding Bellman operator, which is defined as follows,
\begin{align*}
\cT^\pi Q(s, a) = \EE\bigl[ r + \gamma \cdot Q(s', a') \biggiven r\sim R(\cdot \given s, a), s'\sim P(\cdot \given s, a), a'\sim \pi(\cdot \given s') \bigr].
\end{align*}
However, $\cT^\pi Q$ may be not representable by a given function class $\cF$.
Hence, we turn to minimizing a surrogate of the MSBE over $Q\in \cF$, namely the mean-squared projected Bellman error (MSPBE), which is defined as follows,
\begin{align}
\label{eq:mspbe}
\mspbe(Q) = \frac{1}{2} \cdot \EE_{(s, a) \sim \cD} \Bigl[ \bigl( Q(s, a) - \Pi_\cF \cT^\pi  Q(s, a) \bigr)^2 \Bigr],
\end{align}
where $\Pi_\cF $ is the projection onto $\cF$ with respect to the $\cL_2(\cD)$-norm. The global minimizer of the MSPBE is the fixed point solution to the projected Bellman equation $Q = \Pi_\cF \cT^\pi Q$.

In temporal-difference (TD) learning, corresponding to the MSPBE defined in \eqref{eq:mspbe}, we parameterize the Q-function with $\hat Q(\cdot; \theta)$ and update the parameter $\theta$ via stochastic semigradient descent \citep{sutton2018reinforcement},
\begin{align}
\label{eq:sgd}
\theta' = \theta - \epsilon \cdot \bigl(\hat Q(s, a; \theta) - r - \gamma \cdot \hat Q(s', a'; \theta) \bigr) \cdot \nabla_\theta \hat Q(s, a; \theta ),
\end{align}
where $\epsilon >0$ is the stepsize and $(s, a, r, s', a') \sim \tilde\cD$. Here we denote by $\tilde \cD \in \sP(\cS\times \cA\times \RR \times \cS\times \cA)$ the distribution of $(s, a, r, s', a')$, where $(s, a) \sim \cD$, $r\sim R(\cdot \given s, a)$, $s'\sim P(\cdot \given s, a)$, and $a' \sim \pi(\cdot \given s')$.

\subsection{Wasserstein Space}
Let $\Theta \subseteq \RR^D$ be a Polish space. We denote by $\sP_2(\Theta) \subseteq \sP(\Theta)$ the set of probability measures with finite second moments. Then, the Wasserstein-2 distance between $\mu, \nu \in \sP_2(\Theta)$ is defined as follows,
\begin{align}
\label{eq:w2-1}
\cW_2(\mu, \nu) = \inf\Bigl\{ \EE\bigl[\norm{X - Y}^2\bigr]^{1/2} \Biggiven \law (X) = \mu, \law (Y) = \nu  \Bigr\},
\end{align} 
where the infimum is taken over the random variables $X$ and $Y$ on $\Theta$. Here we denote by ${\rm law}(X)$ the distribution of a random variable $X$.
We call $\cM = (\sP_2(\Theta), \cW_2)$ the Wasserstein space, which is an infinite-dimensional manifold \citep{villani2008optimal}. In particular, such a structure allows us to write any tangent vector at $\mu \in \cM$ as $\rho'_0$ for a corresponding curve $\rho: [0, 1] \rightarrow \sP_2(\Theta)$ that satisfies $\rho_0 = \mu$. Here $\rho'_0$ denotes $\partial_t \rho_t\given_{t=0}$. Specifically, under certain regularity conditions, for any curve $\rho: [0, 1] \rightarrow \sP_2(\Theta)$, the continuity equation $\partial_t \rho_t = - \Div(\rho_t v_t)$ corresponds to a vector field $v : [0,1 ]\times \Theta \rightarrow \RR^D$, which endows the infinite-dimensional manifold $\sP_2(\Theta)$ with a weak Riemannian structure in the following sense \citep{villani2008optimal}. Given any tangent vectors $u$ and $\tilde u$ at $\mu\in \cM$ and the corresponding vector fields $v, \tilde v$, which satisfy $u + \Div (\mu v) = 0$ and $\tilde u + \Div(\mu \tilde v) = 0$, respectively, we define the inner product of $u$ and $\tilde u$ as follows,
\begin{align}
\label{eq:w-inner}
\inp{u}{\tilde u}_{\mu} = \int \inp{v}{\tilde v}\, \rd \mu,
\end{align}
which yields a Riemannian metric. Here $\inp{v}{\tilde v} $ is the inner product on $\RR^D$. Such a Riemannian metric further induces a norm $\norm{u}_\mu = \inp{u}{u}_\mu^{1/2}$ for any tangent vector $u\in T_\mu \cM$ at any $\mu\in \cM$, which allows us to write the Wasserstein-2 distance defined in \eqref{eq:w2-1} as follows,
\begin{align}
\label{eq:w2-2}
\cW_2(\mu, \nu) = \inf\Biggl\{ \biggl( \int_0^1 \norm{\rho'_t}^2_{\rho_t} \,\rd t\biggr)^{1/2} \Bigggiven \rho:[0, 1] \rightarrow \cM, \rho_0 = \mu, \rho_1 = \nu\Biggr\}.
\end{align}
Here $\rho'_s$ denotes $\partial_t \rho_t\given_{t=s}$ for any $s\in [0,1]$. In particular, the infimum in \eqref{eq:w2-2} is attained by the geodesic $\tilde \rho: [0,1] \rightarrow \sP_2(\Theta)$ connecting $\mu, \nu \in \cM$. Moreover, the geodesics on $\cM$ are constant-speed, that is, 
\begin{align}
\label{eq:constant-speed}
\norm{\tilde \rho'_t}_{\tilde \rho_t} = \cW_2(\mu, \nu),\quad \forall t\in[0,1].
\end{align}


%% file: algo.tex
\section{Temporal-Difference Learning} \label{sec:td}

For notational simplicity, we write $\RR^d = \RR^{d_1} \times \RR^{d_2}$, $\cX = \cS\times \cA \subseteq \RR^d$, and $x = (s, a) \in \cX$ for any $s\in \cS$ and $a\in \cA$. 

\vskip4pt
\noindent{\bf Parameterization of Q-Function.}
We consider the parameter space $\RR^D$ and parameterize the Q-function with the following two-layer neural network,
\begin{align}
\label{eq:nn-fin}
\hat Q(x; \theta^{(m)})  = \frac{\alpha}{m} \sum_{i=1}^{m} \sigma(x; \theta_i),
\end{align}
where $\theta^{(m)} = (\theta_1, \ldots, \theta_m )\in \RR^{D\times m}$ is the parameter, $m \in \NN_+$ is the width, $\alpha > 0$ is the scaling parameter, and $\sigma: \RR^d \times \RR^D \rightarrow \RR$ is the activation function. Assuming the activation function  in \eqref{eq:nn-fin} takes the form of $\sigma(x; \theta) = b \cdot \tilde \sigma(x; w)$ for $\theta = (w, b)$, we recover the standard form of two-layer neural networks, where $\tilde \sigma$ is the rectified linear unit or the sigmoid function. Such a parameterization is also used in \cite{chizat2018note,mei2019mean,chen2020mean}. For $\{\theta_i\}_{i=1}^m$ independently sampled from a distribution $\rho \in \sP(\RR^D)$, we have the following infinite-width limit of \eqref{eq:nn-fin},
\begin{align}
\label{eq:nn-infty}
Q(x; \rho) = \alpha \cdot \int  \sigma(x; \theta)\,\rd \rho(\theta).
\end{align}
For the empirical distribution $\hat \rho^{(m)} = m^{-1}\cdot \sum_{i=1}^m \delta_{\theta_i}$ corresponding to $\{\theta_i\}_{i=1}^m$, we have $ Q(x; \hat \rho^{(m)}) = \hat Q(x; \theta^{(m)})$.

\vskip4pt
\noindent{\bf TD Dynamics.}
In what follows, we consider the TD dynamics,
\begin{align}
\label{eq:td-fixed}
\theta_i(k+1) = \theta_i(k) - \eta\epsilon \cdot \alpha\cdot \Bigl( \hat Q\bigl(x_k; \theta^{(m)}(k)\bigr) - r_k -  \gamma\cdot \hat Q\bigl(x_k'; \theta^{(m)}(k)\bigr)\Bigr) \cdot\nabla_\theta \sigma\bigl(x_k; \theta_i(k) \bigr),
\end{align}
where $i \in [m]$, $(x_k, r_k, x'_k) \sim \tilde \cD$, and $\epsilon > 0$ is the stepsize with the scaling parameter $\eta > 0$. Without loss of generality, we assume that $(x_k, r_k, x'_k)$ is independently sampled from $\tilde \cD$, while our analysis straightforwardly generalizes to the setting of Markov sampling \citep{bhandari2018finite, zou2019finite, xu2019two}. For an initial distribution $\rho_0 \in\sP(\RR^D)$, we initialize $\{\theta_i\}_{i=1}^m$ as $\theta_i \overset{\iid}{\sim} \rho_0 \ (i\in [m])$. See Algorithm \ref{alg:td} for a detailed description.

\begin{algorithm}
\caption{Temporal-Difference Learning with Two-Layer Neural Network for Policy Evaluation}
\label{alg:td}
\begin{algorithmic}
	\STATE {\bf Initialization:} $\theta_i(0) \overset{\iid}{\sim} \rho_0 \ (i\in [m])$, number of iterations $K = \lfloor T / \epsilon \rfloor$, and policy $\pi$ of interest.
	\FOR{$k = 0, \ldots, K - 1 $}
	 \STATE Sample the state-action pair $(s, a)$ from the stationary distribution $\cD$ of $\pi$, receive the reward $r$, and obtain the subsequent state-action pair $(s', a')$.
	 \STATE Calculate the Bellman residual $\delta = \hat Q(x; \theta^{(m)}(k)) - r - \gamma\cdot  \hat Q(x'; \theta^{(m)}(k))$, where $x = (s, a)$ and $x' = (s', a')$.
	 \STATE Perform the TD update $\theta_i(k+1) \leftarrow \theta_i(k) - \eta \epsilon \cdot \alpha \cdot \delta \cdot \nabla_\theta \sigma(x; \theta_i(k)) \ (i\in [m])$.
	\ENDFOR
	\ENSURE $\{\theta^{(m)}(k)\}_{k=0}^{K-1}$ 
\end{algorithmic}
\end{algorithm}

\vskip4pt

\noindent{\bf Mean-Field Limit.}
Corresponding to $\epsilon\rightarrow 0^+$ and $m\rightarrow \infty$, the continuous-time and infinite-width limit of the TD dynamics in Algorithm \ref{alg:td} is characterized by the following partial differential equation (PDE) with $\rho_0$ as the initial distribution,
\begin{align}
\label{eq:pde-fixed}
\partial_t \rho_t = - \eta \cdot \Div \bigl( \rho_t  \cdot  g(\cdot; \rho_t) \bigr).
\end{align}
Here $g(\cdot; \rho_t): \RR^D \rightarrow \RR^D$ is a vector field, which is defined as follows,
\begin{align}
\label{eq:g-rho}
g(\theta; \rho) = -\alpha \cdot \EE_{(x, r, x') \sim \tilde \cD} \Bigl[ \bigl(Q(x; \rho) - r - \gamma \cdot Q(x'; \rho)\bigr) \cdot \nabla_\theta \sigma(x; \theta) \Bigr].
\end{align}
Note that \eqref{eq:pde-fixed} holds in the sense of distributions \citep{ambrosio2008gradient}.
See \cite{mei2018mean, mei2019mean, araujo2019mean} for the existence, uniqueness, and regularity of the PDE solution $\rho_t$ in \eqref{eq:pde-fixed}. In the sequel, we refer to the continuous-time and infinite-width limit with $\epsilon\rightarrow 0^+$ and $m\rightarrow \infty$ as the mean-field limit.
Let $\hat \rho_k^{(m)} = m^{-1} \cdot \sum_{i=1}^{m} \delta_{\theta_i(k)}$ be the empirical distribution corresponding to $\{\theta_i(k)\}_{i=1}^m$ in \eqref{eq:td-fixed}. The following proposition proves that the PDE solution $\rho_t$ in \eqref{eq:pde-fixed} well approximates the TD dynamics $\theta^{(m)}(k) $ in \eqref{eq:td-fixed}. 
\begin{proposition}[Informal Version of Proposition \ref{prop:discretization}]
	\label{prop:discretization0}
	Let the initial distribution $\rho_0$ be the standard Gaussian distribution $N(0, I_D)$. Under certain regularity conditions, $\hat \rho_{\lfloor t/\epsilon\rfloor}^{(m)} $ weakly converges to $\rho_t$ as $\epsilon\rightarrow 0^+$ and $m\rightarrow \infty$.
\end{proposition} 
The proof of Proposition \ref{prop:discretization0} is based on the  propagation of chaos \citep{sznitman1991topics, mei2018mean, mei2019mean}. 
In contrast to \cite{mei2018mean, mei2019mean}, the PDE in \eqref{eq:pde-fixed} can not be cast as a gradient flow, since there does not exist a corresponding energy functional. Thus, their analysis is not directly applicable to our setting. We defer the detailed discussion on the approximation analysis to \S\ref{sec:discretization}. Proposition \ref{prop:discretization0} allows us to convert the TD dynamics over the finite-dimensional parameter space to its counterpart over the infinite-dimensional Wasserstein space, where the infinitely wide neural network $Q(\cdot;\rho)$ in \eqref{eq:nn-infty} is linear in the distribution $\rho$.
\vskip4pt

\noindent{\bf Feature Representation.} We are interested in the evolution of the feature representation 
\#\label{eq:wfeature}
\Bigl(\nabla_\theta \sigma\bigl(x; \theta_1(k)\bigr)^\top, \ldots, \nabla_\theta \sigma\bigl(x; \theta_m(k)\bigr)^\top\Bigr)^\top \in \RR^{Dm}
\#
corresponding to $\theta^{(m)}(k) = (\theta_1(k), \ldots, \theta_m(k)) \in \RR^{D\times m}$. Such a feature representation is used to analyze the TD dynamics $\theta^{(m)}(k) $ in \eqref{eq:td-fixed} in the NTK regime \citep{cai2019neural}, which corresponds to setting $\alpha = \sqrt{m}$ in \eqref{eq:nn-fin}. Meanwhile, the nonlinear gradient TD dynamics \citep{bhatnagar2009convergent} explicitly uses such a feature representation at each iteration to locally linearize the Q-function. Moreover, up to a rescaling, such a feature representation corresponds to the kernel 
\begin{align*}
\KK(x, x'; \hat \rho_k^{(m)}) = \int \nabla_\theta \sigma(x; \theta)^\top \nabla_\theta \sigma(x'; \theta) \, \rd \hat \rho_k^{(m)}(\theta),
\end{align*}
which by Proposition \ref{prop:discretization0} further induces the kernel
\#\label{eq:wmfkernel}
\KK(x, x'; \rho_t) = \int \nabla_\theta \sigma(x; \theta)^\top \nabla_\theta \sigma(x'; \theta) \, \rd \rho_t(\theta)
\#
at the mean-field limit with $\epsilon\rightarrow 0^+$ and $m\rightarrow \infty$. Such a correspondence allows us to use the PDE solution $\rho_t$ in \eqref{eq:pde-fixed} as a proxy for characterizing the evolution of the feature representation in \eqref{eq:wfeature}.


%% file: result.tex
\section{Main Results}
We first introduce the assumptions for our analysis. In \S\ref{sec:result-rho}, we establish the global optimality and convergence of the PDE solution $\rho_t$ in \eqref{eq:pde-fixed}. In \S\ref{sec:result-theta}, we further invoke Proposition \ref{prop:discretization0} to establish the global optimality and convergence of the TD dynamics $\theta^{(m)}(k) $ in \eqref{eq:td-fixed}.

\begin{assumption}
	\label{asp:data}
	We assume that the state-action pair $x = (s, a) $ satisfies $\norm{x} \le 1$ for any $s \in \cS$ and $a \in \cA$.
\end{assumption}
Assumption \ref{asp:data} can be ensured by normalizing all state-action pairs. Such an assumption is commonly used in the mean-field analysis of neural networks \citep{chizat2018global, mei2018mean, mei2019mean, araujo2019mean, fang2019over, fang2019convex, chen2020mean}. We remark that our analysis straightforwardly generalizes to the setting where $\norm{x} \le C$ for an absolute constant $C > 0$.

\begin{assumption}
	\label{asp:activation}
	We assume that the activation function $\sigma$ in \eqref{eq:nn-fin} satisfies 
	\begin{align}
	\label{eq:activation}
	\bigl| \sigma(x; \theta) \bigr| \le B_0, \quad \bigl\| \nabla_\theta \sigma(x; \theta)\bigr\| \le B_1\cdot \norm{x}, \quad \bigl\|\nabla^2_{\theta \theta} \sigma(x; \theta) \bigr\|_{\rm F} \le B_2 \cdot \norm{x}^2
	\end{align}
	for any $x \in \cX$. Also, we assume that the reward $r$ satisfies $|r| \leq B_r$.
\end{assumption}
Assumption \ref{asp:activation} holds for a broad range of neural networks. For example, let $\theta = (w, b) \in \RR^{D-1} \times \RR$. The activation function
\begin{align}
\label{eq:act1}
\sigma^\dagger(x; \theta) = B_0 \cdot \tanh(b) \cdot \sigmoid(w^\top x)
\end{align}
 satisfies \eqref{eq:activation} in Assumption \ref{asp:activation}. Moreover, the infinitely wide neural network in \eqref{eq:nn-infty} with the activation function $\sigma^\dagger$ in \eqref{eq:act1} induces the following function class,
\begin{align*}
\cF^\dagger = \biggl\{ \int \beta \cdot \sigmoid(w^\top x)\, \rd \mu(w, \beta) \,\bigg|\, \mu \in \sP\bigl(\RR^{D-1} \times [-B_0, B_0]\bigr) \biggr\},
\end{align*}
where $\beta = B_0 \cdot \tanh(b)\in  [-B_0, B_0] $. By the universal approximation theorem \citep{barron1993universal,pinkus1999approximation}, $\cF^\dagger$ captures a rich class of functions. 

\subsection{Global Optimality and Convergence of PDE Solution} \label{sec:result-rho}
Throughout the rest of this paper, we consider the following function class,
\begin{align}
\label{eq:func-class}
\cF = \biggl\{ \int \sigma_{0}(b) \cdot \sigma_{1}(x; w) \,\rd \rho(w, b) \,\bigg| \, \rho \in \sP_2(\RR^{D-1} \times \RR) \biggr\},
\end{align}
which is induced by the infinitely wide neural network in \eqref{eq:nn-infty} with $\theta = (w, b) \in \RR^{D-1} \times \RR$ and the following activation function,
\$
\sigma(x; \theta) = \sigma_{0}(b) \cdot \sigma_{1}(x; w).
\$
We assume that $\sigma_0$ is an odd function, that is, $\sigma_0(b) = - \sigma_0(-b)$, which implies $\int \sigma(x; \theta) \, \rd\rho_0(\theta) = 0$. Note that the set of infinitely wide  neural networks taking the forms of \eqref{eq:nn-infty} is $\alpha \cdot \cF$, which is larger than $\cF$ in \eqref{eq:func-class} by the scaling parameter $\alpha > 0$. Thus, $\alpha$ can be viewed as the degree of ``overrepresentation''. Without loss of generality, we assume that $\cF$ is complete. The following theorem characterizes the global optimality and convergence of the PDE solution $\rho_t$ in \eqref{eq:pde-fixed}.

 
\begin{theorem}\label{th:convergence-fix}
There exists a unique fixed point solution to the projected Bellman equation $Q = \Pi_\cF\cT^\pi Q$, which takes the form of $Q^*(x) = \int \sigma(x; \theta)\,\rd \bar \rho(\theta)$. Also, $Q^*$ is the global minimizer of the MSPBE defined in \eqref{eq:mspbe}. We assume that $D_{\chi^2}(\bar \rho \,\|\, \rho_0) < \infty$ and $\bar \rho(\theta) > 0$ for any $\theta \in \RR^D$. Under Assumptions \ref{asp:data} and \ref{asp:activation}, it holds for $\eta = \alpha^{-2}$ in \eqref{eq:pde-fixed} that
	\begin{align}
	\label{eq:convergence-fix}
	 \inf_{t\in[0, T]}\EE_{x\sim\cD}\Bigl[ \bigl( Q(x; \rho_t) - Q^*(x)\bigr)^2 \Bigr] \le \frac{D_{\chi^2}(\bar \rho \,\|\, \rho_0)}{2(1-\gamma)\cdot T} + \frac{C_*}{(1-\gamma) \cdot \alpha} ,
	\end{align}
	where $C_*>0$ is a constant that depends on $D_{\chi^2}(\bar \rho \,\|\, \rho_0)$, $B_1$, $B_2$, and $B_r$. 
\end{theorem}
Theorem \ref{th:convergence-fix} proves that the optimality gap $\EE_{x\sim\cD}[ ( Q(x; \rho_t) - Q^*(x))^2 ]$ decays to zero at a sublinear rate up to the error of $O(\alpha^{-1})$, where $\alpha > 0$ is the scaling parameter in \eqref{eq:nn-fin}. Varying $\alpha$ leads to a tradeoff between such an error of $O(\alpha^{-1})$ and the deviation of $\rho_t$ from $\rho_0$. Specifically, in \S\ref{sec:proof} we prove that $\rho_t$ deviates from $\rho_0$ by the divergence $D_{\chi^2}(\rho_t \,\|\, \rho_0) \leq O(\alpha^{-2})$. Hence, a smaller $\alpha$ allows $\rho_t$ to move further away from $\rho_0$, inducing a feature representation that is more different from the initial one \citep{fang2019over,fang2019convex}. See \eqref{eq:wfeature}-\eqref{eq:wmfkernel} for the correspondence of $\rho_t$ with the feature representation and the kernel that it induces. On the other hand, a smaller $\alpha$ yields a larger error of $O(\alpha^{-1})$ in \eqref{eq:convergence-fix} of Theorem \ref{th:convergence-fix}. In contrast, the NTK regime \citep{cai2019neural}, which corresponds to setting $\alpha = \sqrt{m}$ in \eqref{eq:nn-fin}, only allows $\rho_t$ to deviate from $\rho_0$ by the divergence $D_{\chi^2}(\rho_t \,\|\, \rho_0) \leq O(m^{-1}) = o(1)$. In other words, the NTK regime fails to induce a feature representation that is significantly different from the initial one. In summary, our analysis goes beyond the NTK regime, which allows us to characterize the evolution of the feature representation towards the (near-)optimal one.

\subsection{Global Optimality and Convergence of TD Dynamics} \label{sec:result-theta}

As a result of Proposition \ref{prop:discretization0}, we establish the following lemma, which characterizes the error of approximating the optimality gap in Theorem \ref{th:convergence-fix} by that of the TD dynamics $\theta^{(m)}(k) $ in \eqref{eq:td-fixed}.
\begin{lemma}
	\label{lem:opt-dis}
	Let $B$ be a constant that depends on $\alpha$, $\eta$, $\gamma$, $B_0$, $B_1$, and $B_2$.
	Under Assumptions \ref{asp:data} and \ref{asp:activation}, it holds for any $k\le T/\epsilon \ (k\in \NN)$ that
	\begin{align*}
	&\EE_{x\sim \cD} \biggl[ \Bigl( \hat Q\bigl(x; \theta^{(m)}(k)\bigr) - Q^*(x) \Bigr)^2 \biggr] \nonumber \\
	&\quad \le \EE_{x\sim\cD} \Bigl[ \bigl( Q(x; \rho_{k\epsilon}) - Q^*(x)\bigr)^2 \Bigr] + B\cdot e^{BT} \cdot \Bigl( \sqrt{m^{-1} \cdot \log(m/\delta)}  + \sqrt{\epsilon\cdot \bigl(D + \log(m/\delta)\bigr)}\Bigr)
	\end{align*}
	with probability at least $1-\delta$.
\end{lemma}
\begin{proof}
	 See \S\ref{sec:pf-opt-dis} for a detailed proof.
\end{proof}

Based on Theorem \ref{th:convergence-fix} and Lemma \ref{lem:opt-dis}, we establish the following corollary, which characterizes the global optimality and convergence of the TD dynamics $\theta^{(m)}(k) $ in \eqref{eq:td-fixed}.
\begin{corollary}
	\label{cor:conv-td}
	Under the same conditions of Theorem \ref{th:convergence-fix}, it holds with probability at least $1-\delta$ that
	\begin{align}
	\label{eq:conv-td}
	\min_{\substack{k \le T/\epsilon \\ (k \in \NN)}} \EE_{x\sim \cD} \biggl[ \Bigl( \hat Q\bigl(x; \theta^{(m)}(k) \bigr) - Q^*(x) \Bigr)^2 \biggr] \le \frac{D_{\chi^2}(\bar \rho \,\|\, \rho_0)}{2(1-\gamma)\cdot T} + \frac{C_*}{(1-\gamma) \cdot \alpha} + \Delta(\epsilon, m,\delta, T),
	\end{align}
	where $C_*>0$ is the constant in \eqref{eq:convergence-fix} of Theorem \ref{th:convergence-fix} and $\Delta(\epsilon, m, \delta, T) > 0$ is an error term such that
	\begin{align*}
	\lim_{m\rightarrow \infty} \lim_{\epsilon\rightarrow 0^+} \Delta(\epsilon, m, \delta, T) = 0.
	\end{align*}
\end{corollary}
\begin{proof}
	Combining Theorem \ref{th:convergence-fix} and Lemma \ref{lem:opt-dis} implies Corollary \ref{cor:conv-td}.
\end{proof}
In \eqref{eq:conv-td} of Corollary \ref{cor:conv-td}, the error term $\Delta(\epsilon, m, \delta, T)$ characterizes the error of approximating the TD dynamics $\theta^{(m)}(k) $ in \eqref{eq:td-fixed} using the PDE solution $\rho_t$ in \eqref{eq:pde-fixed}. In particular, such an error vanishes at the mean-field limit. 


%% file: proof.tex

\section{Proof of Main Results} \label{sec:proof}
We first introduce two technical lemmas. Recall that $\cF$ is defined in \eqref{eq:func-class}, $Q(x; \rho)$ is defined in \eqref{eq:nn-infty}, and $g(\theta; \rho)$ is defined in \eqref{eq:g-rho}.
\begin{lemma}
   \label{lem:rho-star}
   There exists a unique fixed point solution to the projected Bellman equation $Q = \Pi_\cF \cT^\pi Q$, which takes the form of $Q^*(x) = \int \sigma(x; \theta)\,\rd \bar \rho(\theta)$. Also, there exists $\rho^* \in \sP_2(\RR^D)$ that satisfies the following properties,
   \begin{itemize}
   	\item[(i)] $Q(x; \rho^*) = Q^*(x)$ for any $x\in \cX$,
   	\item[(ii)] $g(\cdot; \rho^*) = 0$ for $\bar \rho$-a.e., and
   	\item[(iii)] $\cW_2(\rho^*, \rho_0) \le \alpha^{-1}\cdot \bar D$, where  $\bar D = D_{\chi^2}(\bar \rho\,\|\,\rho_0)^{1/2}$.
   \end{itemize}
\end{lemma}
\begin{proof}
	See \S\ref{sec:pf-rho-star} for a detailed proof.
\end{proof}   

Lemma \ref{lem:rho-star} establishes the existence of the fixed point solution $Q^*$ to the projected Bellman equation $Q = \Pi_\cF\cT^\pi Q$. Furthermore, such a fixed point solution $Q^*$ can be parameterized with  the infinitely wide neural network $Q(\cdot; \rho^*)$ in \eqref{eq:nn-infty}. Meanwhile, the Wasserstein-2 distance between $\rho^*$ and the initial distribution $\rho_0$ is upper bounded by $O(\alpha^{-1})$. Based on the existence of $Q^*$ and the property of $\rho^*$ in Lemma \ref{lem:rho-star}, we establish the following lemma that characterizes the evolution of $ \cW_2(\rho_t, \rho^*) $, where $\rho_t$ is the PDE solution in \eqref{eq:pde-fixed}.


\begin{figure}

	\tikzset{every picture/.style={line width=0.75pt}} 
	\centering
	\begin{tikzpicture}[x=0.75pt,y=0.75pt,yscale=-1,xscale=1]
	
	\draw  [draw opacity=0] (233.93,187.57) .. controls (240.4,170.49) and (255.28,154.51) .. (278.32,142.16) .. controls (333.75,112.47) and (416.66,114.4) .. (469.64,145.29) -- (372.6,208.64) -- cycle ; \draw   (233.93,187.57) .. controls (240.4,170.49) and (255.28,154.51) .. (278.32,142.16) .. controls (333.75,112.47) and (416.66,114.4) .. (469.64,145.29) ;
	\draw [color={rgb, 255:red, 3; green, 0; blue, 0 }  ,draw opacity=1 ][fill={rgb, 255:red, 0; green, 0; blue, 0 }  ,fill opacity=1 ] [dash pattern={on 0.84pt off 2.51pt}]  (280.66,140.89) -- (481,156.2) ;
	\draw [shift={(380.83,148.54)}, rotate = 184.37] [color={rgb, 255:red, 3; green, 0; blue, 0 }  ,draw opacity=1 ][line width=0.75]    (10.93,-3.29) .. controls (6.95,-1.4) and (3.31,-0.3) .. (0,0) .. controls (3.31,0.3) and (6.95,1.4) .. (10.93,3.29)   ;
	\draw    (480.41,156.38) -- (481,156.2) ;
	\draw  [dash pattern={on 0.84pt off 2.51pt}]  (280.66,140.89) -- (340.13,101.7) ;
	\draw [shift={(341.8,100.6)}, rotate = 506.62] [color={rgb, 255:red, 0; green, 0; blue, 0 }  ][line width=0.75]    (10.93,-3.29) .. controls (6.95,-1.4) and (3.31,-0.3) .. (0,0) .. controls (3.31,0.3) and (6.95,1.4) .. (10.93,3.29)   ;
	
	\draw (271.6,145) node [anchor=north west][inner sep=0.75pt]   [align=left] {$\displaystyle \rho _{t}$};
	\draw (339.48,87.85) node   [align=left] {$\displaystyle g( \cdot ; \rho _{t})$};
	\draw (368.6,153) node [anchor=north west][inner sep=0.75pt]   [align=left] {$\displaystyle v$};
	\draw (475.4,156.3) node [anchor=north west][inner sep=0.75pt]   [align=left] {$\displaystyle \rho ^{*}$};
	
	\draw [fill={rgb, 255:red, 0; green, 0; blue, 0 }  ,fill opacity=1 ][line width=0.75]   (280.75, 140.89) circle [x radius= 2, y radius= 2]   ;
	\draw [fill={rgb, 255:red, 0; green, 0; blue, 0 }  ,fill opacity=1 ][line width=0.75]   (481, 156.2) circle [x radius= 2, y radius= 2]   ;
	\draw [fill={rgb, 255:red, 0; green, 0; blue, 0 }  ,fill opacity=1 ][line width=0.75]   (280.66, 140.89) circle [x radius= 2, y radius= 2]   ;
	\end{tikzpicture}
	
	\caption{We illustrate the first variation formula $\frac{\rd \cW_2(\rho_t, \rho^*)^2}{2} = - \inp{g(\cdot; \rho_t)}{v}_{\rho_t}$, where $v$ is the vector field corresponding to the geodesic that connects $\rho_t$ and $\rho^*$. See Lemma \ref{lem:diff} for details.}
	\label{fig:descent}
\end{figure}
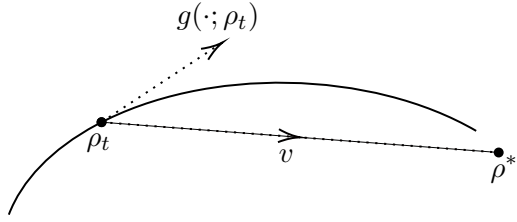

\begin{lemma}
	\label{lem:descent}
	 We assume that $\cW_2(\rho_t, \rho^*) \le 2\cW_2(\rho_0, \rho^*)$, $D_{\chi^2}(\bar \rho \,\|\, \rho_0) < \infty$, and $\bar \rho(\theta) > 0$ for any $\theta \in \RR^D$. Under Assumptions \ref{asp:data} and \ref{asp:activation}, it holds that 
	\begin{align}
	\label{eq:lem-descent}
	\frac{\rd }{\rd t}\frac{\cW_2(\rho_t, \rho^*)^2}{2}  \le - (1-\gamma) \cdot \eta\cdot \EE_{x \sim \cD}\Bigl[ \bigl( Q(x; \rho_t) - Q^*(x)\bigr)^2 \Bigr] + C_* \cdot \alpha^{-1} \cdot \eta , 
	\end{align}
	where $C_* > 0$ is a constant depending on $D_{\chi^2}(\bar \rho\,\|\,\rho_0)$, $B_1$, $B_2$, and $B_r$.
\end{lemma}
\begin{proof}
	See \S\ref{sec:pf-descent} for a detailed proof.
\end{proof}
The proof of Lemma \ref{lem:descent} is based on the first variation formula of the Wasserstein-2 distance (Lemma \ref{lem:diff}), which is illustrated in Figure \ref{fig:descent}, and the one-point monotonicity of $g(\cdot; \beta_t)$ along a curve $\beta$ on the Wasserstein space (Lemma \ref{lem:td-q}). When the right-hand side of \eqref{eq:lem-descent} is nonpositive, Lemma \ref{lem:descent} characterizes the decay of $\cW_2(\rho_t, \rho^*)$. We are now ready to present the proof of Theorem \ref{th:convergence-fix}.


	\begin{proof}
	We use a continuous counterpart of the induction argument. We define
	\begin{align}
	\label{eq:t-star1}
	t^* = \inf\biggl\{ \tau \in \RR_+ \bigggiven  \EE_{x \sim \cD}\Bigl[ (1-\gamma)\cdot \bigl( Q(x; \rho_\tau) - Q^*(x)\bigr)^2 \Bigr ] <  C_* \cdot \alpha^{-1}\biggr\}.
	\end{align}
	In other words, the right-hand side of \eqref{eq:lem-descent} in Lemma \ref{lem:descent} is nonpositive for any $t\le t^*$, that is,
	\begin{align}
	\label{eq:pf01}
	- (1-\gamma) \cdot \EE_{x \sim \cD}\Bigl[ \bigl( Q(x; \rho_t) - Q^*(x)\bigr)^2 \Bigr] + C_* \cdot \alpha^{-1} \le 0.
	\end{align}
	Also, we define 
	\begin{align}
	\label{eq:t-star2}
	t_* = \inf\bigl\{ \tau \in \RR_+ \biggiven \cW_2 (\rho_\tau, \rho^*) > 2 \cW_2 (\rho_0, \rho^*) \bigr\}.
	\end{align}
    In other words, \eqref{eq:lem-descent} of Lemma \ref{lem:descent} holds for any $t \le t_*$. Thus, for any $0 \le t \le \min\{t^*, t_*\}$, it holds that $\frac{\rd}{\rd t} \frac{\cW_2(\rho_t, \rho_*)^2}{2} \le 0$.  Figure \ref{fig:axis} illustrates the definition of $t^*$ and $t_*$ in \eqref{eq:t-star1} and \eqref{eq:t-star2}, respectively. 
        
    \tikzset{every picture/.style={line width=0.75pt}} 
    
    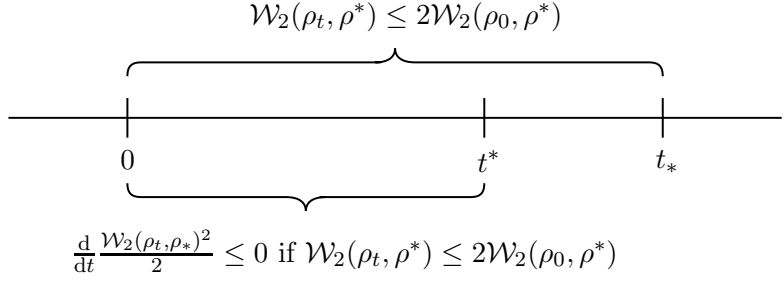
\begin{figure}
    \centering
    \begin{tikzpicture}[x=0.75pt,y=0.75pt,yscale=-1,xscale=1]

    
    \draw    (150,260) -- (540,260) ;
    \draw    (210,250) -- (210,270) ;
    \draw    (390,250) -- (390,270) ;
    \draw    (480,250) -- (480,270) ;
    \draw   (210,293) .. controls (210,298) and (212,300) .. (217,300) -- (290,300) .. controls (297,300) and (300,303) .. (300,307) .. controls (300,303) and (303,300) .. (310,300)(307,300) -- (383,300) .. controls (388,300) and (390,298) .. (390,293) ;
    \draw   (480,239) .. controls (480,234) and (478,232) .. (473,232) -- (355,232) .. controls (348,232) and (345,229) .. (345,225) .. controls (345,229) and (342,232) .. (335,232)(338,232) -- (217,232) .. controls (212,232) and (210,234) .. (210,239) ;
    
    \draw (205,275) node [anchor=north west][inner sep=0.75pt]    {$0$};
    \draw (385,275) node [anchor=north west][inner sep=0.75pt]   [align=left] {$\displaystyle t^{*}$};
    \draw (475,275) node [anchor=north west][inner sep=0.75pt]   [align=left] {$\displaystyle t_{*}$};
    \draw (270,200) node [anchor=north west][inner sep=0.75pt]   [align=left] {$\cW_2(\rho_t, \rho^*)\le 2 \cW_2(\rho_0, \rho^*)$};
    \draw (180,315) node [anchor=north west][inner sep=0.75pt]   [align=left] { $\frac{\rd}{\rd t} \frac{\cW_2(\rho_t, \rho_*)^2}{2} \le 0$ if $\cW_2(\rho_t, \rho^*)\le 2 \cW_2(\rho_0, \rho^*)$};
    \end{tikzpicture}
    \caption{For any $ 0 \le t \le \min\{t^*, t_*\}$, \eqref{eq:lem-descent} of Lemma \ref{lem:descent} holds and $\frac{\rd}{\rd t} \frac{\cW_2(\rho_t, \rho_*)^2}{2} \le 0$. }
    \label{fig:axis}
    \end{figure}
    
    We now prove that $t_* \ge t^*$ by contradiction. By the continuity of $\cW_2(\rho_t, \rho^*)^2$ with respect to $t$ \citep{ambrosio2008gradient}, it holds that $t_* > 0$, since $\cW_2(\rho_0, \rho^*) < 2 \cW_2(\rho_0, \rho^*)$. For the sake of contradiction, we assume that $t_* < t^*$, by \eqref{eq:lem-descent} of Lemma \ref{lem:descent} and \eqref{eq:pf01}, it holds for any $0 \leq t \leq t_*$ that 
	\begin{align*}
	\frac{\rd }{\rd t}\frac{\cW_2(\rho_t, \rho^*)^2}{2}  \le 0,
	\end{align*}
	which implies that $\cW_2(\rho_t, \rho^*) \le \cW_2(\rho_0, \rho^*)$ for any $0 \leq t \leq t_*$. This contradicts the definition of $t_*$ in \eqref{eq:t-star2}. Thus, it holds that $t_* \ge t^*$, which implies that \eqref{eq:lem-descent} of Lemma \ref{lem:descent} holds for any $0 \leq t\le t^*$.
	
	If $t^* \le T$, \eqref{eq:pf01} implies Theorem \ref{th:convergence-fix}. 
	If $t^* > T$, by \eqref{eq:lem-descent} of Lemma \ref{lem:descent}, it holds for any $0 \leq t\leq T$ that
	\begin{align*}
	\frac{\rd }{\rd t}\frac{\cW_2(\rho_t, \rho^*)^2}{2}  \le - (1-\gamma) \cdot \eta\cdot \EE_{x \sim \cD}\Bigl[ \bigl( Q(x; \rho_t) - Q^*(x)\bigr)^2 \Bigr] + C_* \cdot \alpha^{-1} \cdot \eta \le 0,
	\end{align*}
	which further implies that
	\begin{align}
	\label{eq:pf2}
	\EE_{x \sim \cD}\Bigl[ \bigl( Q(x; \rho_t) - Q^*(x)\bigr)^2 \Bigr] \le -(1-\gamma)^{-1} \cdot \eta^{-1} \cdot \frac{\rd }{\rd t} \frac{\cW_2(\rho_t, \rho^*)^2}{2}  + C_*\cdot (1-\gamma)^{-1} \cdot  \alpha^{-1}.
	\end{align}
	Upon telescoping \eqref{eq:pf2} and setting $\eta = \alpha^{-2}$, we obtain that
	\begin{align*}
	&\inf_{t\in[0, T]}\EE_\cD\Bigl[ \bigl( Q(x; \rho_t) - Q^*(x)\bigr)^2 \Bigr] \nonumber\\ &\quad \le T^{-1}\cdot \int_{0}^{T}\EE_{x \sim \cD}\Bigl[ \bigl( Q(x; \rho_t) - Q^*(x)\bigr)^2 \Bigr] \,\rd t \nonumber\\
	&\quad \le 1/2\cdot (1-\gamma)^{-1} \cdot \eta^{-1} \cdot T^{-1} \cdot \cW_2(\rho_0, \rho^*)^2 + C_*\cdot (1-\gamma)^{-1} \cdot  \alpha^{-1} \nonumber \\
	&\quad \le 1/2\cdot (1-\gamma)^{-1} \cdot \bar D^2 \cdot T^{-1} + C_*\cdot (1-\gamma)^{-1}\cdot  \alpha^{-1},
	\end{align*}
	where the last inequality follows from the fact that $\eta = \alpha^{-2}$ and (iii) of Lemma \ref{lem:rho-star}.
	Thus, we complete the proof of Theorem \ref{th:convergence-fix}.		
\end{proof}

%% file: extension.tex
\section{Extension to Q-Learning and Policy Gradient}
\label{sec:extension}
In this section, we extend our analysis of TD to Q-learning and policy gradient. In \S\ref{sec:q-learn}, we introduce Q-learning and its mean-field limit. In \S\ref{sec:q-conv}, we establish the global optimality and convergence of Q-learning. In \S\ref{sec:softpolicy}, we further extend our analysis to soft Q-learning, which is equivalent to policy gradient. 
\subsection{Q-Learning}\label{sec:q-learn}
Q-learning aims to solve the following projected Bellman optimality equation,
\begin{align}
\label{eq:bellman-opt}
Q = \Pi_\cF \cT^* Q.
\end{align}
Here $\cT^*$ is the Bellman optimality operator, which is defined as follows,
\begin{align*}
\cT^* Q(s, a) = \EE\bigl[r + \gamma \cdot \max_{\underline a\in \cA} Q(s', \underline a) \biggiven r\sim R(\cdot \given s, a), s'\sim P(\cdot \given s, a) \bigr].
\end{align*}
When $\Pi_\cF$ is the identity mapping, the fixed point solution to \eqref{eq:bellman-opt} is the Q-function $Q^{\pi^*}$ of the optimal policy $\pi^*$, which maximizes the expected total reward $J(\pi)$ defined in \eqref{eq:def-j} \citep{sutton2018reinforcement}.  We consider the parameterization of the Q-function in \eqref{eq:nn-fin} and update the parameter $\theta^{(m)}$ as follows,
\begin{align}
\label{eq:q-sgd}
&\theta_i(k+1) \\
&\quad= \theta_i(k) - \eta\epsilon \cdot \alpha \cdot \Bigl( \hat Q\bigl(s_k, a_k; \theta^{(m)}(k)\bigr) - r_k - \gamma \cdot \max_{\underline a\in \cA}\hat Q\bigl(s'_k, \underline a; \theta^{(m)}(k) \bigr)  \Bigr) \cdot \nabla_\theta \sigma \bigl(s_k, a_k; \theta_i(k)\bigr), \notag
\end{align}
where $i \in [m]$, $(s_k, a_k)$ is sampled from the stationary distribution $\cD_{\rE} \in \sP(\cS\times \cA)$ of an exploration policy $\pi_\rE$, $r_k \sim R(\cdot\given s_k, a_k)$ is the reward, and $s'_k \sim P(\cdot\given s_k, a_k)$ is the subsequent state. For notational simplicity, we denote by $\tilde \cD_\rE \in \sP(\cS\times \cA \times \RR \times \cS)$ the distribution of $(s_k, a_k, r_k, s'_k)$. For an initial distribution $\nu_0 \in \sP(\RR^D)$, we initialize $\{\theta_i\}_{i=1}^m$ as $\theta_i \overset{\iid}{\sim} \rho_0 \ (i\in [m])$. See Algorithm \ref{alg:q-learning} for a detailed description.
\begin{algorithm}
	\caption{Q-Learning with Two-Layer Neural Network for Policy Improvement}
	\label{alg:q-learning}
	\begin{algorithmic}
		\STATE {\bf Initialization.} $\theta_i(0) \overset{\iid}{\sim} \nu_0 \ (i\in [m])$, number of iterations $K = \lfloor T / \epsilon \rfloor$, and exploration policy $\pi_\rE$.
		\FOR{$k = 0, \ldots, K - 1 $}
		\STATE Sample the state-action pair $(s, a)$ from the stationary distribution $\cD_\rE$ of $\pi_\rE$, receive the reward $r$, and obtain the subsequent state $s'$. 
		\STATE Calculate the Bellman residual $\delta = \hat Q(x; \theta^{(m)}(k)) - r - \gamma \cdot \hat Q(x'; \theta^{(m)}(k))$, where $x = (s, a)$ and $x' = (s', \argmax_{\underline a\in \cA} \hat Q(s', \underline a; \theta^{(m)}(k)))$.
		\STATE Perform the Q-learning update $\theta_i(k+1) \leftarrow \theta_i(k) - \eta \epsilon \cdot \alpha \cdot \delta \cdot \nabla_\theta \sigma(x; \theta_i(k)) \ (i\in [m])$.
		\ENDFOR
		\ENSURE $\{ \theta^{(m)}(k) \}_{k=0}^{K-1} $
	\end{algorithmic}
\end{algorithm}

\vskip4pt

\noindent{\bf Mean-Field Limit.}
Corresponding to $\epsilon \rightarrow 0^+$ and $m\rightarrow \infty$,
the mean-field limit of the Q-learning dynamics in Algorithm \ref{alg:q-learning} is characterized by the following PDE with $\nu_0$ as the initial distribution,
\begin{align}
\label{eq:q-pde}
\partial_t \nu_t = -\eta \cdot \Div\bigl(\nu_t \cdot h(\cdot; \nu_t)\bigr).
\end{align}
Here $h(\cdot; \nu_t) : \RR^D \rightarrow \RR^D$ is a vector field, which is defined as follows,
\begin{align}
\label{eq:q-grad}
h(\theta; \nu) = - \alpha \cdot \EE_{(s, a, r, s') \sim \tilde \cD_\rE}\Bigl[ \bigl( Q(s, a; \nu) - r - \gamma \cdot \max_{\underline a \in \cA} Q(s', \underline a; \nu)\bigr) \cdot \nabla_\theta \sigma(s, a; \theta) \Bigr].
\end{align}
In parallel to Proposition \ref{prop:discretization0}, the empirical distribution $\hat \nu^{(m)}_k = m^{-1} \cdot \sum_{i=1}^m \delta_{\theta_i(k)}$ weakly converges to $\nu_{k\epsilon}$ as $\epsilon \rightarrow 0^+$ and $m\rightarrow \infty$. 

\subsection{Global Optimality and Convergence of Q-Learning} \label{sec:q-conv}
The $\max$ operator in the Bellman optimality operator $\cT^*$ makes the analysis of Q-learning more challenging than that of TD. Correspondingly, we lay out an extra regularity condition on the exploration policy $\pi_\rE$. Recall that the function class $\cF$ is defined in \eqref{eq:func-class}.

\begin{assumption}
	\label{asp:exp}
	We assume for an absolute constant $\kappa > 0$ and any $Q^1, Q^2 \in \cF$ that
	\begin{align*}
    \EE_{(s, a) \sim \cD_\rE} \Bigl[ \bigl( Q^1(s, a) - Q^2(s, a) \bigr)^2 \Bigr] \ge (\gamma + \kappa)^2 \cdot \EE_{(s, a) \sim \cD_\rE}\Bigl[ \bigl( \max_{\underline a\in \cA}Q^1(s, \underline a) - \max_{\underline a\in \cA}Q^2(s, \underline a) \bigr)^2 \Bigr].
	\end{align*}
\end{assumption}
Although Assumption \ref{asp:exp} is strong, we are not aware of any weaker regularity condition in the literature, even in the linear setting \citep{melo2008analysis,zou2019finite, chen2019performance} and the NTK regime \citep{cai2019neural}. Let the initial distribution $\nu_0$ be the standard Gaussian distribution $N(0, I_D)$.
In parallel to Theorem \ref{th:convergence-fix}, we establish the following theorem, which characterizes the global optimality and convergence of Q-learning. Recall that we write $\cX = \cS\times \cA$ and $x = (s, a) \in \cX$. Also, $\nu_t$ is the PDE solution in \eqref{eq:q-pde}, while $\theta^{(m)}(k)$ is the Q-learning dynamics in \eqref{eq:q-sgd}.

\begin{theorem}\label{th:q-convergence}
	There exists a unique fixed point solution to the projected Bellman optimality equation $Q = \Pi_\cF\cT^* Q$, which takes the form of 
	 $Q^\dagger(x) = \int \sigma(x; \theta) \,\rd \bar \nu(\theta)$. We assume that $D_{\chi^2}(\bar \nu \,\|\, \nu_0) < \infty$ and $\bar \nu(\theta) > 0$ for any $\theta \in \RR^D$.
	Under Assumptions \ref{asp:data}, \ref{asp:activation}, and \ref{asp:exp}, it holds for $\eta = \alpha^{-2}$ that
	\begin{align}
	\label{eq:q-convergence}
	\inf_{t\in[0, T]}\EE_{x \sim \cD_\rE}\Bigl[ \bigl( Q(x; \nu_t) - Q^\dagger (x)\bigr)^2 \Bigr] \le \frac{ (\kappa+\gamma) \cdot  D_{\chi^2}(\bar \nu \,\|\, \nu_0)}{2\kappa \cdot T} + \frac{(\kappa+\gamma) \cdot C_*}{\kappa \cdot \alpha},
	\end{align}
	where $C_* > 0$ is a constant depending on $D_{\chi^2}(\bar \nu \,\|\,\nu_0)$, $B_1$, $B_2$, and $B_r$. Moreover, it holds with probability at least $1-\delta$ that
	\begin{align}
	\label{eq:conv-q}
	&\min_{\substack{k \le T/\epsilon \\ (k \in \NN)}} \EE_{x \sim \cD_\rE} \biggl[ \Bigl( \hat Q\bigl(x; \theta^{(m)}(k) \bigr) - Q^\dagger(x) \Bigr)^2 \biggr] \notag\\
	&\quad\le \frac{(\kappa + \gamma) \cdot D_{\chi^2}(\bar \nu \,\|\, \nu_0)}{2\kappa \cdot T} + \frac{(\kappa + \gamma) \cdot C_*}{\kappa \cdot \alpha} + \Delta(\epsilon, m, \delta, T),
	\end{align}
	where $\Delta(\epsilon, m, \delta, T) > 0$ is an error term such that
	\$
	\lim_{ m\rightarrow \infty} \lim_{\epsilon\rightarrow 0^+}\Delta(\epsilon, m, \delta, T) = 0.
	\$
\end{theorem}
\begin{proof}
	See \S\ref{sec:pf-q-convergenve} for a detailed proof.
\end{proof}
Theorem \ref{th:q-convergence} proves that the optimality gap $\EE_{x\sim \cD_\rE} [ ( Q(x; \nu_t) - Q^\dagger(x))^2 ] $ decays to zero at a sublinear rate up to the error of $O(\alpha^{-1})$, where $\alpha > 0$ is the scaling parameter in \eqref{eq:nn-fin}. In parallel to Theorem \ref{th:convergence-fix}, varying $\alpha$ leads to a tradeoff between such an error of $O(\alpha^{-1})$ and the deviation of $\nu_t$ from $\nu_0$. 
Moreover, based on the counterparts of Proposition \ref{prop:discretization0} and Lemma \ref{lem:opt-dis}, Theorem \ref{th:q-convergence} gives the global optimality and convergence of the Q-learning dynamics $\theta^{(m)}(k)$ in \eqref{eq:q-sgd}, which is in parallel to Corollary \ref{cor:conv-td}. 


\subsection{Soft Q-Learning and Policy Gradient}\label{sec:softpolicy}
Theorem \ref{th:q-convergence} straightforwardly generalizes to soft Q-learning, where the max operator is replaced by the softmax operator. Specifically, we define the soft Bellman optimality operator as follows,
\begin{align*}
\cT_\beta Q (s, a) = \EE\bigl[ r + \gamma \cdot {\softmax_{\underline a \in \cA}}^\beta Q(s', \underline a) \biggiven r \sim R(\cdot\given s, a), s' \sim P(\cdot\given s, a) \bigr],
\end{align*}
where the softmax operator is defined as follows,
\begin{align*}
{\softmax_{\underline a \in \cA}}^\beta Q(s, \underline a) = \beta \cdot \log \EE_{\underline a \sim \bar \pi(\cdot\given s)}\Bigl[ \exp \bigl( \beta^{-1} \cdot Q(s, \underline a) \bigr)\Bigr].
\end{align*}
Here $\bar \pi(\cdot\given s)$ is the uniform policy.
Soft Q-learning aims to find the fixed point solution to the projected soft Bellman optimality equation $ Q = \Pi_\cF \cT_\beta Q$. 
In parallel to the Q-learning dynamics in \eqref{eq:q-sgd}, we consider the following soft Q-learning dynamics,
\begin{align}
\label{eq:soft-sgd}
&\theta_i(k+1) \\
&\quad= \theta_i(k) - \eta\epsilon \cdot \alpha \cdot \Bigl( \hat Q\bigl(s_k, a_k; \theta^{(m)}(k)\bigr) - r_k - \gamma \cdot {\softmax_{\underline a \in \cA}}^\beta \hat Q\bigl(s'_k, \underline a; \theta^{(m)}(k) \bigr)  \Bigr) \cdot \nabla_\theta \sigma \bigl(s_k, a_k; \theta_i(k)\bigr),\notag
\end{align}
whose mean-field limit is characterized by the following PDE,
\begin{align}
\label{eq:soft-pde}
\partial_t \nu_t = -\eta \cdot \Div\bigl(\nu_t \cdot h(\cdot; \nu_t)\bigr).
\end{align}
In parallel to \eqref{eq:q-grad}, $h(\cdot; \nu_t) : \RR^D \rightarrow \RR^D$ is a vector field, which is defined as follows,
\begin{align*}
h(\theta; \nu) = - \alpha \cdot \EE_{(s, a, r, s') \sim \tilde \cD_\rE}\Bigl[ \bigl( Q(s, a; \nu) - r - \gamma \cdot {\softmax_{\underline a \in \cA}}^\beta Q(s', \underline a; \nu)\bigr) \cdot \nabla_\theta \sigma(s, a; \theta) \Bigr].
\end{align*}
In parallel to Assumption \ref{asp:exp}, we lay out the following regularity condition.
\begin{assumption}
	\label{asp:exp-soft}
	We assume for an absolute constant $\kappa > 0$ and any $\nu^1, \nu^2 \in \sP(\RR^D)$ that
	\begin{align*}
	&\EE_{(s, a) \sim \cD_\rE} \Bigl[ \bigl( Q(s, a; \nu^1) - Q(s, a; \nu^2) \bigr)^2 \Bigr] \\ &\quad \ge (\gamma + \kappa)^2 \cdot \EE_{(s, a) \sim \cD_\rE}\Bigl[ \bigl( {\softmax_{\underline a \in \cA}}^\beta Q(s, \underline a; \nu^1) - {\softmax_{\underline a \in \cA}}^\beta Q(s, \underline a; \nu^2) \bigr)^2 \Bigr].
	\end{align*}
\end{assumption}
The following proposition parallels Theorem \ref{th:q-convergence}, which characterizes the global optimality and convergence of soft Q-learning.
Recall that $\nu_t$ is the PDE solution in \eqref{eq:soft-pde} and $\theta^{(m)}(k)$ is the soft Q-learning dynamics in \eqref{eq:soft-sgd}.
\begin{proposition}\label{prop:soft-convergence}
	There exists a unique fixed point solution to the projected soft Bellman optimality equation $Q = \Pi_\cF\cT_\beta Q$, which takes the form of 
	$Q^\ddagger(x) = \int \sigma(x; \theta) \,\rd \underline \nu(\theta)$. We assume that $D_{\chi^2}(\underline \nu \,\|\, \nu_0) < \infty$ and $\underline \nu(\theta) > 0$ for any $\theta \in \RR^D$.
	Under Assumptions \ref{asp:data}, \ref{asp:activation}, and \ref{asp:exp-soft}, it holds for $\eta = \alpha^{-2}$ that
	\begin{align*}
	\inf_{t\in[0, T]}\EE_{x \sim \cD_\rE}\Bigl[ \bigl( Q(x; \nu_t) - Q^\ddagger (x)\bigr)^2 \Bigr] \le \frac{ (\kappa+\gamma) \cdot  D_{\chi^2}(\underline \nu \,\|\, \nu_0)}{2\kappa \cdot T} + \frac{(\kappa+\gamma) \cdot C_*}{\kappa \cdot \alpha},
	\end{align*}
	where $C_* > 0$ is a constant depending on $D_{\chi^2}(\underline \nu \,\|\,\nu_0)$, $B_1$, $B_2$, and $B_r$. Moreover, it holds with probability at least $1-\delta$ that
	\begin{align*}
	\min_{\substack{k \le T/\epsilon \\ (k \in \NN)}} \EE_{x\sim \cD_\rE} \biggl[ \Bigl( \hat Q\bigl(x; \theta^{(m)}(k) \bigr) - Q^\ddagger(x) \Bigr)^2 \biggr] \le \frac{(\kappa + \gamma) \cdot D_{\chi^2}(\underline \nu \,\|\, \nu_0)}{2\kappa \cdot T} + \frac{(\kappa + \gamma) \cdot C_*}{\kappa \cdot \alpha} + \Delta(\epsilon, m, \delta, T),
	\end{align*}
	where $\Delta(\epsilon, m, \delta, T) > 0$ is an error term such that
	\$
	\lim_{ m\rightarrow \infty} \lim_{\epsilon\rightarrow 0^+}\Delta(\epsilon, m, \delta, T) = 0.
	\$
\end{proposition}
\begin{proof}
	Replacing the max operator by the softmax operator in the proof of Theorem \ref{th:q-convergence}  implies Proposition \ref{prop:soft-convergence}. 
\end{proof}
Moreover, soft Q-learning is equivalent to a variant of policy gradient \citep{o2016combining,schulman2017equivalence,nachum2017bridging,haarnoja2017reinforcement}. Hence, Proposition \ref{prop:soft-convergence} also characterizes the global optimality and convergence of such a variant of policy gradient. 

%
%

%% file: appendix.tex
\newpage
\appendix
\section{Proofs for \S\ref{sec:proof}-\ref{sec:extension}} 
For notational simplicity, we denote by $\EE_{\cD}$ the expectation with respect to $x \sim \cD$ and $\EE_{\tilde \cD}$ the expectation with respect to $(x, r, x') \sim \tilde \cD$. Also, with a slight abuse of notations, we write $\theta^{(m)} = \{\theta_i\}_{i=1}^m$.
\subsection{Proof of Lemma \ref{lem:rho-star}} \label{sec:pf-rho-star}
\begin{proof}
	
	\noindent{\bf Existence and uniqueness of $Q^*$.}
	To establish the existence of the fixed point solution $Q^*$ to the projected Bellman equation $ Q = \Pi_\cF\cT^\pi Q $, it suffices to show that $\Pi_\cF\cT^\pi: \cF\rightarrow \cF$ is a contraction mapping. It holds for any $Q^1, Q^2 \in \cF$ that
	\begin{align*}
	\norm{\Pi_\cF\cT^\pi Q^1 - \Pi_\cF\cT^\pi Q^2}_{\cL_2(\cD)}^2 &\le  \gamma^2\cdot \EE_{\tilde \cD}\Bigl[ \bigl(Q^1(x') - Q^2(x')  \bigr)^2 \Bigr] \nonumber \\
	&= \gamma^2\cdot \norm[\big]{Q^1 - Q^2}_{\cL_2(\cD)}^2,
	\end{align*}
	where the last equality follows from the fact that $\cD$ is the stationary distribution. Thus, $\Pi_\cF\cT^\pi: \cF\rightarrow \cF$ is a contraction mapping. Note that $\cF$ is complete.
	Following from the Banach fixed point theorem \citep{conway2019course}, there exists a unique $Q^* \in \cF$ that solves the projected Bellman equation $Q = \Pi_\cF\cT^\pi Q$. Moreover, by the definition of $\cF$ in \eqref{eq:func-class}, there exists $\bar\rho \in \sP_2(\RR^D)$ such that
	\begin{align*}
	Q^*(x) = \int \sigma(x; \theta) \, \rd\bar\rho( \theta).
	\end{align*}
	
	\vskip4pt
	
	\noindent{\bf Proof of (i) in Lemma \ref{lem:rho-star}.} 
	We define
	\begin{align}
	\label{eq:rho-bar}
	\rho^* = \rho_0 + \alpha^{-1} \cdot (\bar \rho - \rho_0).
	\end{align}
	By the definition of $Q(\cdot ; \rho)$ in \eqref{eq:nn-infty} and the fact that $Q(x; \rho_0) = 0$, we have that $Q(x; \rho^*) = Q^*(x)$, which completes the proof of (i) in Lemma \ref{lem:rho-star}.
	
	\vskip4pt
	
	\noindent{\bf Proof of (ii) in Lemma \ref{lem:rho-star}.}
	For (ii) of Lemma \ref{lem:rho-star}, note that $Q(\cdot; \rho^*) = \Pi_\cF\cT^\pi Q(\cdot; \rho^*)$. Thus, we have that
	\begin{align*}
	\inp[\big]{ Q(\cdot; \rho^*) - \cT^\pi Q(\cdot; \rho^*) }{f(\cdot ) - Q(\cdot; \rho^*)}_{\cD} \ge 0, \quad \forall f\in \cF,
	\end{align*}
	which further implies that
	\begin{align}
	\label{eq:pf-rh2}
	\EE_{\tilde \cD}\Bigl[ \bigl( Q(x; \rho^*) - r - \gamma\cdot  Q(x'; \rho^*) \bigr) \cdot \int \sigma(x; \theta) \,\rd ( \rho - \bar \rho)(\theta) \Bigr] \ge 0, \quad \forall \rho\in \sP_2(\RR^D).
	\end{align}
	Let $\rho = (\textrm{id} + h \cdot v)_\sharp \bar \rho$ for a sufficiently small scaling parameter $h\in \RR_+$ and any Lipschitz-continuous mapping $v : \RR^D \rightarrow \RR^D$. Then, following from \eqref{eq:pf-rh2}, we have that
	\begin{align}
	\label{eq:pf-rh3}
	\int \EE_{\tilde \cD}\biggl[ \bigl( Q(x; \rho^*) - r - \gamma \cdot Q(x'; \rho^*) \bigr) \cdot \Bigl(\sigma\bigl(x; \theta+ h\cdot v(\theta)\bigr) - \sigma(x;\theta) \Bigr) \biggr]  \,\rd \bar \rho(\theta) \ge 0
	\end{align}
	for any $ v : \RR^D \rightarrow \RR^D$.
	Dividing the both sides of \eqref{eq:pf-rh3} by $h$ and letting $h\rightarrow 0^+$, we have for any $ v: \RR^D \rightarrow \RR^D$ that
	\begin{align*}
	0 &\le \int \EE_{\tilde \cD}\Bigl[ \bigl( Q(x; \rho^*) - r - \gamma \cdot Q(x'; \rho^*) \bigr) \cdot \inp[\big]{\nabla_\theta \sigma(x; \theta)}{v(\theta)} \Bigr]  \,\rd \bar \rho(\theta) \nonumber\\
	& = -\alpha^{-1} \cdot \int \inp[\big]{g(\theta; \rho^*)}{v(\theta)}\, \rd \bar \rho(\theta),
	\end{align*}
	where the equality follows from the definition of $g$ in \eqref{eq:g-rho}.
	Thus, we have that $g(\theta; \rho^*) = 0$ for $\bar \rho$-a.e., which completes the proof of (ii) in Lemma \ref{lem:rho-star}.
	
	\vskip4pt
	
	\noindent{\bf Proof of (iii) in Lemma \ref{lem:rho-star}.}
	Following from the definition of $\rho^*$ in \eqref{eq:rho-bar}, we have that
	\begin{align*}
	&D_{\chi^2}(\rho^* \,\|\, \rho_0) \nonumber \\
	&\quad = \int  \biggl( \frac{\rho^*(\theta)}{\rho_0(\theta)} - 1 \biggr)^2 \,\rd \rho_0(\theta) = \int \biggl( \frac{(1-\alpha^{-1}) \cdot \rho_0(\theta) + \alpha^{-1}\cdot \bar \rho(\theta)}{\rho_0(\theta)} - 1\biggr) \,\rd\rho_0(\theta)   = \alpha^{-2} \cdot \bar D^2,
	\end{align*}
	where $\bar{D} = D_{\chi^2}(\bar{\rho} \,\|\, \rho_0)^{1/2}$. By Lemma \ref{lem:talagrand}, we have that
	\begin{align*}
	\cW_2(\rho^*, \rho_0) \le D_{\rm KL}(\rho^*\,\|\,\rho_0)^{1/2} \le  D_{\chi^2}(\rho^* \,\|\, \rho_0)^{1/2} \le \alpha^{-1} \cdot \bar D,
	\end{align*}
	which completes the proof of (iii) in Lemma \ref{lem:rho-star}.
\end{proof}

\subsection{Proof of Lemma \ref{lem:descent}}\label{sec:pf-descent}
We first introduce the following lemmas. The first lemma establishes the one-point monotonicity of $g(\cdot; \beta_t)$ along a curve $\beta: [0,1 ] \rightarrow \sP_2(\RR^D)$ on the Wasserstein space.

\begin{lemma}
	\label{lem:td-q}
	Let $\beta: [0, 1] \rightarrow \sP_2(\RR^D)$ be a curve such that $\partial_t \beta_t = -\Div(\beta_t \cdot v_t)$ for a vector field $v$. We have that
	\begin{align*}
	\inp[\big]{\partial_t g(\cdot; \beta_t)}{v_t}_{\beta_t} \le -(1-\gamma) \cdot \EE_\cD\Bigl[ \bigl( \partial_t Q(x; \beta_t) \bigr)^2 \Bigr].
	\end{align*}
	Furthermore, we have that
	\begin{align}
	\label{eq:td-q1}
	\int_{0}^{1}\inp[\big]{\partial_s g(\cdot; \beta_s)}{v_s}_{\beta_s} \, \rd s \le - (1-\gamma) \cdot \EE_\cD\Bigl[ \bigl( Q(x; \beta_0) - Q(x; \beta_1)\bigr)^2 \Bigr].
	\end{align}
\end{lemma}

\begin{proof}
	Following from the definition of $g$ in \eqref{eq:g-rho}, we have that
	\begin{align*}
	\partial_t g(\theta; \beta_t) = -\alpha \cdot \EE_{\tilde \cD} \Bigl[ \partial_t \bigl(Q(x; \beta_t) - \gamma \cdot Q(x'; \beta_t)\bigr) \cdot \nabla_\theta \sigma(x; \theta) \Bigr].
	\end{align*}
	Thus, following from integration by parts and the continuity equation $\partial_t \beta_t =- \Div(\beta_t \cdot v_t)$, we have that
	\begin{align}
	\label{eq:pf-td-q1}
	\inp[\big]{\partial_t g(\cdot; \beta_t)}{ v_t}_{\beta_t} 
	&= -\int \inp[\bigg]{\alpha \cdot \EE_{\tilde \cD} \Bigl[ \partial_t \bigl(Q(x; \beta_t) - \gamma \cdot Q(x'; \beta_t)\bigr) \cdot \nabla_\theta \sigma(x; \theta) \Bigr]}{v_t(\theta) \cdot \beta_t(\theta)}\,\rd \theta \nonumber \\
	&= - \int \alpha \cdot \EE_{\tilde \cD} \Bigl[ \partial_t \bigl(Q(x; \beta_t) - \gamma \cdot Q(x'; \beta_t)\bigr) \cdot \sigma(x; \theta)\Bigr] \cdot \partial_t \beta_t(\theta) \, \rd \theta \nonumber \\
	&= - \EE_{\tilde \cD} \Bigl[ \partial_t \bigl(Q(x; \beta_t) - \gamma \cdot Q(x'; \beta_t)\bigr) \cdot \partial_t Q(x; \beta_t)\Bigr],
	\end{align}
	where the last equality follows from the definition of $Q$ in \eqref{eq:nn-infty}. Applying the Cauchy-Schwartz inequality to \eqref{eq:pf-td-q1}, we have that
	\begin{align}
	\label{eq:pf-td-q2}
	\inp[\big]{\partial_t g(\cdot; \beta_t)}{ v_t}_{\beta_t} &= - \EE_{\tilde \cD} \Bigl[ \bigl(\partial_t Q(x; \beta_t)\bigr)^2\Bigr] + \gamma \cdot \EE_{\tilde \cD}\bigl[ \partial_t Q(x'; \beta_t) \cdot \partial_t Q(x; \beta_t)\bigr] \nonumber \\
	&\le - \EE_{\tilde \cD} \Bigl[ \bigl(\partial_t Q(x; \beta_t)\bigr)^2\Bigr] + \gamma \cdot \EE_{\tilde \cD} \Bigl[ \bigl(\partial_t Q(x; \beta_t)\bigr)^2\Bigr]^{1/2} \cdot \EE_{\tilde \cD} \Bigl[ \bigl(\partial_t Q(x'; \beta_t)\bigr)^2\Bigr]^{1/2} \nonumber \\
	& = -(1-\gamma) \cdot \EE_{\cD} \Bigl[ \bigl(\partial_t Q(x; \beta_t)\bigr)^2\Bigr],
	\end{align}
	where the last equality follows from the fact that the marginal distributions of $\tilde \cD$ with respect to $x$ and $x'$ are $\cD$, since $\cD$ is the stationary distribution.
	Furthermore, we have that
	\begin{align*}
	\int_0^1 \inp[\big]{\partial_s g(\cdot; \beta_s)}{ v_s}_{\beta_s} \, \rd s &\le - (1-\gamma) \cdot \int_0^1 \EE_\cD\Bigl[ \bigl( \partial_s Q(x; \beta_s) \bigr)^2 \Bigr] \, \rd s \\
	& \le - (1-\gamma) \cdot \EE_\cD\biggl[ \Bigl( \int_0^1 \partial_s Q(x; \beta_s) \, \rd s\Bigr)^2 \biggr] \\
	&= - (1-\gamma) \cdot \EE_\cD\Bigl[ \bigl( Q(x; \beta_1) - Q(x; \beta_0)\bigr)^2 \Bigr],
	\end{align*}
	which completes the proof of Lemma \ref{lem:td-q}.
\end{proof}

The following lemma upper bounds the norms of $Q$ and $\nabla_\theta g$.
\begin{lemma}
	\label{lem:boundness}
	Under Assumptions \ref{asp:data} and \ref{asp:activation}, it holds for any $\rho\in \sP_2(\RR^D)$ that
	\begin{align}
	\label{eq:lem-boundness1}
	\sup_{x\in \cX} \bigl| Q(x; \rho)\bigr| &\le \alpha\cdot \min \bigl\{B_1 \cdot \cW_2(\rho, \rho_0),\, B_0\bigr\}, \\
	\label{eq:lem-boundness2}
	\sup_{\theta\in \RR^D}\norm[\big]{\nabla_\theta g(\theta; \rho)}_{\rm F} &\le \alpha \cdot B_2 \cdot \min\bigl\{ 2\alpha \cdot B_1 \cdot \cW_2(\rho, \rho_0) + B_r,\, 2\alpha\cdot B_0 + B_r \bigr\}.
	\end{align}
\end{lemma}
\begin{proof}
	We introduce the Wasserstein-1 distance, which is defined as 
	\begin{align*}
	\cW_1(\mu^1, \mu^2) = \inf\Bigl\{ \EE\bigl[\norm{X-Y}\bigr] \Biggiven {\rm law}(X) = \mu^1, {\rm law}(Y) = \mu^2 \Bigr\}
	\end{align*}
	for any $\mu^1, \mu^2 \in\sP(\RR^D)$ with finite first moments. Thus, we have that $\cW_1(\mu^1, \mu^2) \le \cW_2(\mu^1, \mu^2)$. The Wasserstein-1 distance has the following dual representation \citep{ambrosio2008gradient},
	\begin{align}
	\label{eq:w-1-dual}
	\cW_1(\mu^1, \mu^2) = \sup\biggl\{ \int f(x) \,\rd (\mu^1-\mu^2)(x) \bigggiven {\rm continuous}~f: \RR^D \rightarrow \RR, \lip(f) \le 1 \biggr\}.
	\end{align}
	Following from Assumptions \ref{asp:data} and \ref{asp:activation}, we have that $\norm{\nabla_\theta\sigma(x; \theta)} \le B_1$ for any $x\in \cX$ and $\theta\in \RR^D$, which implies that $\lip(\sigma(x; \cdot)/B_1) \le 1$ for any $x \in \cX$. Note that $Q(x; \rho_0) = 0$ for any $x \in \cX$. Thus, by \eqref{eq:w-1-dual} we have for any  $ \rho\in \sP_2(\RR^D) $ and $x\in \cX$ that
	\begin{align}
	\label{eq:pf-bo1}
	\bigl| Q(x; \rho) \bigr| &= \alpha\cdot \biggl| \int \sigma(x; \theta) \cdot  \,\rd (\rho - \rho_0)(\theta) \biggr|  \le \alpha\cdot B_1 \cdot \cW_1(\rho, \rho_0) \le \alpha \cdot B_1 \cdot \cW_2(\rho, \rho_0).
	\end{align}
	Meanwhile, following from Assumptions \ref{asp:data} and \ref{asp:activation}, we have for any $x\in \cX$ and $\rho\in \sP_2(\RR^D)$ that
	\begin{align}
	\label{eq:pf-bo3}
	\bigl| Q(x; \rho) \bigr| = \alpha\cdot \biggl| \int \sigma(x; \theta)   \,\rd \rho(\theta )\biggr| \le \alpha\cdot B_0.
	\end{align}
	Combining \eqref{eq:pf-bo1} and \eqref{eq:pf-bo3}, we have for any $\rho \in \sP_2(\RR^D)$ that
	\begin{align}\label{eq:pf-bo4}
	\sup_{x\in \cX} \bigl| Q(x; \rho)\bigr| &\le \alpha\cdot \min \bigl\{B_1 \cdot \cW_2(\rho, \rho_0),\, B_0\bigr\},
	\end{align}
	which completes the proof of \eqref{eq:lem-boundness1} in Lemma \ref{lem:boundness}.
	Following from the definition of $g$ in \eqref{eq:g-rho}, we have for any $x\in \cX$ and $\rho\in \sP_2(\RR^D)$ that
	\begin{align*}
	\bigl\|\nabla_\theta g(\theta; \rho) \bigr\|_{\rm F} &\le  \alpha \cdot \EE_{\tilde \cD}\Bigl[ \bigl| Q(x; \rho) - r - \gamma\cdot Q(x'; \rho)\bigr|\cdot \norm[\big]{\nabla_{\theta\theta}^2 \sigma(x; \theta)}_{\rm F} \Bigr] \nonumber\\
	&\le \alpha  \cdot \min\bigl\{ 2\alpha \cdot B_1 \cdot \cW_2(\rho, \rho_0) + B_r,\, 2\alpha\cdot B_0 + B_r \bigr\} \cdot B_2.
	\end{align*} 
	Here the last inequality follows from \eqref{eq:pf-bo4} and the fact that $ \norm{\nabla_{\theta\theta}^2 \sigma(x; \theta)}_{\rm F} \le B_2 $ for any $x\in \cX$ and $\rho\in \sP_2(\RR^D)$, which follows from Assumptions \ref{asp:data} and \ref{asp:activation}. Thus, we complete the proof of Lemma \ref{lem:boundness}.
\end{proof}

We are now ready to present the proof of Lemma \ref{lem:descent}.
\begin{proof}
	Recall that $\rho_t$ is the PDE solution in \eqref{eq:pde-fixed}, that is,
	\begin{align*}
	\partial_t \rho_t = - \eta \cdot \Div \bigl(\rho_t \cdot g(\cdot; \rho_t) \bigr),
	\end{align*}
	where 
	\begin{align*}
	g(\theta; \rho) = -\alpha \cdot \EE_{\tilde \cD} \Bigl[ \bigl(Q(x; \rho) - r - \gamma\cdot Q(x'; \rho)\bigr) \cdot \nabla_\theta \sigma(x; \theta) \Bigr].
	\end{align*}
	We fix a $t \in [0, T]$. We denote by $\beta:[0, 1]\rightarrow \sP_2(\RR^D)$ the geodesic connecting $\rho_t$ and $\rho^*$. Specifically, $\beta$ satisfies that $\beta'_s = - \Div(\beta_s \cdot  v_s)$ for a vector field $v$. Following from Lemma \ref{lem:diff}, we have that
	\begin{align}
	\label{eq:pf-ds1}
	\frac{\rd }{\rd t}\frac{\cW_2(\rho_t, \rho^*)^2}{2}  &= -\eta\cdot \inp[\big]{ g(\cdot ;\rho_t) }{ v_0 }_{\rho_t} \nonumber \\
	& = \eta\cdot \int_{0}^{1} \partial_s \inp[\big]{ g(\cdot ;\beta_s) }{ v_s }_{\beta_s} \,\rd s - \eta\cdot \inp[\big]{ g(\cdot ;\rho^*) }{ v_1 }_{\rho^*} \nonumber \\
	& = \eta \cdot \underbrace{\int_0^1 \inp[\big]{ \partial_s g(\cdot; \beta_s) }{ v_s}_{\beta_s} \,\rd s}_{\displaystyle {\rm (i)}} + \eta \cdot \underbrace{ \int_0^1 \int \inp[\big]{ g(\theta; \beta_s) }{\partial_s(v_s \cdot \beta_s)(\theta) } \,\rd \theta\,\rd s}_{\displaystyle {\rm (ii)}},
	\end{align}
	where the last equality follows from (ii) of Lemma \ref{lem:rho-star}. 
	
	For term (i) of \eqref{eq:pf-ds1}, following from \eqref{eq:td-q1} of Lemma \ref{lem:td-q}, we have that
	\begin{align}
	\label{eq:pf-ds2}
	\int_{0}^{1}\inp[\big]{\partial_s g(\cdot; \beta_s)}{v_s}_{\beta_s} \, \rd s &\le - (1-\gamma) \cdot \EE_\cD\Bigl[ \bigl( Q(x; \beta_0) - Q(x; \beta_1)\bigr)^2 \Bigr] \nonumber\\
	&= - (1-\gamma) \cdot \EE_\cD\Bigl[ \bigl( Q(x; \rho_t) - Q^*(x)\bigr)^2 \Bigr].
	\end{align}
	For term (ii) of \eqref{eq:pf-ds2}, we have that
	\begin{align*}
	\int \Bigl| \inp[\big]{ g(\theta; \beta_s) }{\partial_s(v_s \cdot  \beta_s)(\theta) } \Bigr| \,\rd \theta &= \int \Bigl| \inp[\big]{\nabla_\theta g(\theta; \beta_s)}{ \beta_s(\theta)\cdot v_s(\theta) \otimes v_s(\theta) }\Bigr| \,\rd \theta \\
	&\le \sup_{\theta\in \RR^D} \norm[\big]{\nabla_\theta g(\theta; \beta_s)}_{\rm F} \cdot \norm{v_s}_{\beta_s}^2,
	\end{align*}
	where the equality follows from integration by parts and Lemma \ref{lem:euler}. Since $\beta$ is the geodesic connecting $\rho_t$ and $\rho^*$, \eqref{eq:constant-speed} implies that $\norm{v_s}_{\beta_s}^2 = \cW_2(\beta_0, \beta_1)^2 = \cW_2(\rho_t, \rho^*)^2$ for any $s\in [0,1]$.
	Applying \eqref{eq:lem-boundness2} of Lemma \ref{lem:boundness}, we have that
	\begin{align}\label{eq:pf-ds3}
	\int \Bigl| \inp[\big]{ g(\theta; \beta_s) }{\partial_s(v_s \cdot \beta_s)(\theta) } \Bigr| \,\rd \theta
	&\le \alpha \cdot B_2 \cdot \bigl( 2\alpha \cdot B_1 \cdot \cW_2(\rho_t, \rho_0) + B_r \bigr) \cdot \cW_2(\rho_t, \rho^*)^2 \nonumber\\
	&\le 4\alpha \cdot B_2 \cdot \bigl( 6\alpha \cdot B_1 \cdot \cW_2(\rho_0, \rho^*) + B_r \bigr) \cdot \cW_2(\rho_0, \rho^*)^2 ,
	\end{align}
	where the last inequality follows from the condition of Lemma \ref{lem:descent} that $\cW_2(\rho_t, \rho^*) \le 2\cW_2(\rho_0, \rho^*)$ and the fact that $\cW_2(\rho_t, \rho_0) \le \cW_2(\rho_t, \rho^*) + \cW_2(\rho_0, \rho^*)$. Then, applying (iii) of Lemma \ref{lem:rho-star} to \eqref{eq:pf-ds3}, we have that
	\begin{align}
	\label{eq:pf-ds4}
	\int_0^1 \int \Bigl| \inp[\big]{ g(\theta; \beta_s) }{\partial_s(v_s \cdot \beta_s)(\theta) } \Bigr| \,\rd \theta \,\rd s &\le 4\alpha^{-1} \cdot B_2 \cdot \bar D^2 \cdot ( 6  B_1 \cdot \bar D+ B_r ) \nonumber \\
	&=C_* \cdot \alpha^{-1},
	\end{align} 
	where $C_*>0$ is a constant depending on $\bar D$, $B_1$, $B_2$, and $B_r$.
	
	Finally, plugging \eqref{eq:pf-ds2} and \eqref{eq:pf-ds4} into \eqref{eq:pf-ds1}, we have that
	\begin{align*}
	\frac{\rd }{\rd t}\frac{\cW_2(\rho_t, \rho^*)^2}{2} \le - (1-\gamma) \cdot \eta\cdot \EE_\cD\Bigl[ \bigl( Q(x; \rho_t) - Q^*(x)\bigr)^2 \Bigr] + C_* \cdot \alpha^{-1} \cdot \eta, 
	\end{align*}
	which completes the proof of Lemma \ref{lem:descent}.
\end{proof}

\subsection{Proof of Theorem \ref{th:q-convergence}} \label{sec:pf-q-convergenve}
\begin{proof}
	In parallel to the proof of Lemma \ref{lem:rho-star} in \S\ref{sec:pf-rho-star}, to establish the existence and uniqueness of the fixed point solution to the projected Bellman optimality equation $Q = \Pi_\cF \cT^* Q$, it suffices to show that $\Pi_\cF\cT^*:\cF\rightarrow \cF$ is a contraction mapping. In particular, it holds for any $Q^1, Q^2 \in \cF$ that
	\begin{align*}
	\norm{\Pi_\cF \cT^*Q^1 - \Pi_\cF \cT^* Q^2}^2_{\cL_2(\cD_{\rE})} &\le \gamma^2 \cdot \EE_{\tilde \cD_\rE}\Bigl[\bigl( \max_{a\in \cA} Q^{1}(s', \underline a) - \max_{a\in \cA} Q^{2}(s', \underline a) \bigr)^2 \Bigr] \\
	& = \gamma^2 \cdot \EE_{\cD_\rE}\Bigl[\bigl( \max_{a\in \cA} Q^{1}(s, \underline a) - \max_{a\in \cA} Q^{2}(s, \underline a) \bigr)^2 \Bigr] \\
	&\le \frac{\gamma^2}{(\gamma+\kappa)^2} \cdot  \EE_{\cD_\rE}\Bigl[\bigl( Q^{1}(s, a) -  Q^{2}(s, a) \bigr)^2 \Bigr],
	\end{align*}
	where the equality follows from the fact that $\cD_{\rE}$ is the stationary distribution and the last inequality follows from Assumption \ref{asp:exp}. Thus, $\Pi_\cF\cT^*:\cF \rightarrow \cF$ is a contraction mapping. Following from the Banach fixed point theorem \citep{conway2019course}, there exists a unique fixed point solution $Q^\dagger\in \cF$ to the projected Bellman optimality equation $Q = \Pi_\cF \cT^* Q$. Moreover, in parallel to the proof of Lemma \ref{lem:rho-star} in \S\ref{sec:pf-rho-star}, there exists $\nu^\dagger \in \sP_2(\RR^D)$ such that $Q(x; \nu^\dagger) = Q^\dagger(x)$, $h(x; \nu^\dagger) = 0$, and $\cW_2(\nu^\dagger, \nu_0) \le \alpha^{-1} \cdot \bar D$, where $\bar D = D_{\chi^2}(\bar \nu \,\|\, \nu_0)^{1/2}$. 
	
	For notational simplicity, we define $Q^\cA(x) = \max_{\underline a\in \cA} Q(s, \underline a)$.
	In parallel to \eqref{eq:pf-ds1} in the proof of Lemma \ref{lem:descent} in \S\ref{sec:pf-descent}, we have that
	\begin{align}
	\label{eq:pf-qd1}
	\frac{\rd }{\rd t}\frac{\cW_2(\nu_t, \nu^\dagger)^2}{2} = \eta \cdot \underbrace{\int_0^1 \inp[\big]{ \partial_s h(\cdot; \beta_s) }{ v_s}_{\beta_s} \,\rd s}_{\displaystyle {\rm (i)}} + \eta \cdot \underbrace{ \int_0^1 \int \inp[\big]{ h(\theta; \beta_s) }{\partial_s(v_s \cdot \beta_s)(\theta) } \,\rd \theta \,\rd s}_{\displaystyle {\rm (ii)}},
	\end{align}
	where $\beta: [0, 1] \rightarrow \sP_2(\RR^D)$ is the geodesic connecting $\nu_t$ and $\nu^\dagger$ with $\partial_s \beta_s = -\Div(\beta_s\cdot  v_s)$.
	
	\vskip4pt
	
	\noindent{\bf Upper bounding term (i) of \eqref{eq:pf-qd1}.} In parallel to \eqref{eq:pf-td-q1} and \eqref{eq:pf-td-q2} in the proof of Lemma \ref{lem:td-q}, we have that
	\begin{align}
	\label{eq:pf-qd2}
	\inp[\big]{\partial_s h(\cdot; \beta_s)}{ v_s}_{\beta_s} &= - \EE_{\tilde \cD_\rE} \Bigl[ \partial_s \bigl(Q(x; \beta_s) - \gamma \cdot Q^\cA(x'; \beta_s)\bigr) \cdot \partial_s Q(x; \beta_s)\Bigr] \\
	&\le -\EE_{\cD_\rE} \Bigl[ \bigl(\partial_s Q(x; \beta_s)\bigr)^2 \Bigr] + \gamma \cdot \EE_{\cD_\rE} \Bigl[ \bigl(\partial_s Q(x; \beta_s)\bigr)^2 \Bigr]^{1/2} \cdot \EE_{\cD_\rE} \Bigl[ \bigl(\partial_s Q^\cA(x; \beta_s)\bigr)^2 \Bigr]^{1/2}. \nonumber
	\end{align}
	For the second term on the right-hand side of \eqref{eq:pf-qd2}, we have that
	\begin{align}
	\label{eq:pf-qd3}
	\EE_{\cD_\rE} \Bigl[ \bigl(\partial_s Q^\cA(x; \beta_s)\bigr)^2 \Bigr] 
	& = \lim_{u\rightarrow 0}  \EE_{\cD_{\rE}}\biggl[\Bigl( u^{-1} \cdot \bigl( Q^\cA(x; \beta_{s+u}) - Q^\cA(x; \beta_{s}) \bigr)\Bigr)^2 \biggr] \nonumber \\
	& \le (\gamma + \kappa)^{-2} \cdot  \lim_{u\rightarrow 0} u^{-2} \cdot \EE_{\cD_\rE}\Bigl[ \bigl( Q(x; \beta_{s+u}) - Q(x; \beta_{s}) \bigr)^2 \Bigr] \nonumber \\
	& = (\gamma + \kappa)^{-2} \cdot \EE_{\cD_\rE}\Bigl[ \bigl( \partial_s Q(x; \beta_{s}) \bigr)^2 \Bigr],
	\end{align}
	where the inequality follows from Assumption \ref{asp:exp} and the fact that $Q(\cdot; \nu) \in \alpha \cdot \cF$. Plugging \eqref{eq:pf-qd3} into \eqref{eq:pf-qd2}, we have that
	\begin{align*}
	\inp[\big]{\partial_s h(\cdot; \beta_s)}{ v_s}_{\beta_s} \le -\frac{\kappa}{\gamma + \kappa} \cdot \EE_{\cD_\rE}\Bigl[ \bigl( \partial_s Q(x; \beta_{s}) \bigr)^2 \Bigr],
	\end{align*}
	which further implies that
	\begin{align}
	\label{eq:pf-qd4}
	\int_0^1 \inp[\big]{\partial_s h(\cdot; \beta_s)}{ v_s}_{\beta_s} \, \rd s &\le -\frac{\kappa}{\gamma + \kappa} \cdot \int_0^1 \EE_{\cD_\rE}\Bigl[ \bigl( \partial_s Q(x; \beta_{s}) \bigr)^2 \Bigr] \,\rd s \nonumber \\
	& \le -\frac{\kappa}{\gamma + \kappa} \cdot \EE_{\cD_\rE}\biggl[ \Bigl( \int_0^1 \partial_s Q(x; \beta_{s}) \,\rd s \Bigr)^2 \biggr] \nonumber\\
	&= -\frac{\kappa}{\gamma + \kappa} \cdot \EE_{\cD_\rE}\Bigl[ \bigl( Q(x; \nu_t) - Q(x; \nu^\dagger) \bigr)^2 \Bigr].
	\end{align}
	\noindent{\bf Upper bounding term (ii) of \eqref{eq:pf-qd1}.} In parallel to the proof of Lemma \ref{lem:boundness} in \S\ref{sec:pf-descent}, noting that $|Q^\cA(x; \nu)| \le \sup_{x\in \cX} |Q(x; \nu)|$ for any $\nu \in \sP_2(\RR^D)$,  we have that
	\begin{align*}
	\norm[\big]{\nabla_\theta h(\theta;\nu_t)}_{\rm F} \le \alpha \cdot B_2 \cdot \bigl( 2\alpha \cdot B_1 \cdot \cW_2(\nu_t, \nu_0) + B_r \bigr).
	\end{align*}
	In parallel to \eqref{eq:pf-ds3} and \eqref{eq:pf-ds4}, we have that
	\begin{align}\label{eq:pf-qd5}
	\int_0^1 \int \Bigl| \inp[\big]{ h(\theta; \beta_s) }{\partial_s(v_s \cdot  \beta_s)(\theta) }\Bigr| \,\rd \theta\,\rd s \le  C_* \cdot \alpha^{-1},
	\end{align}
	where $C_*>0$ is a constant that depends on $\bar D$, $B_1$, $B_2$, and $B_r$.
	
	Plugging \eqref{eq:pf-qd4} and \eqref{eq:pf-qd5} into \eqref{eq:pf-qd1}, we have that
	\begin{align*}
	\frac{\rd }{\rd t} \frac{\cW_2(\nu_t, \nu^\dagger)^2}{2} \le -\frac{\eta \cdot \kappa}{\gamma + \kappa} \cdot \EE_{\cD_\rE}\Bigl[ \bigl( Q(x; \nu_t) - Q(x; \nu^\dagger) \bigr)^2 \Bigr] + C_* \cdot \eta \cdot \alpha^{-1}.
	\end{align*}
	Thus, in parallel to the proof of Theorem \ref{th:convergence-fix} in \S\ref{sec:proof}, we have that
	\begin{align*}
	\inf_{t\in[0, T]}\EE_\cD\Bigl[ \bigl( Q(x; \nu_t) - Q^\dagger(x)\bigr)^2 \Bigr] \le \frac{ (\kappa+\gamma) \cdot  D_{\chi^2}(\bar \nu \,\|\, \nu_0)}{2\kappa \cdot T} + C_*\cdot \alpha^{-1} \cdot \frac{\kappa + \gamma}{\kappa},
	\end{align*}
	which completes the proof of \eqref{eq:q-convergence} in Theorem \ref{th:q-convergence}. Meanwhile, in parallel to the proof of Lemma \ref{lem:opt-dis} in \S\ref{sec:pf-opt-dis}, we  upper bound the error of approximating $\hat \nu_k$ by $\nu_{k\epsilon}$, which further implies \eqref{eq:conv-q} of Theorem \ref{th:q-convergence}.
\end{proof}

\section{Mean-Field Limit of Neural Networks}
\label{sec:discretization}
In this section, we  prove Proposition \ref{prop:discretization0}, whose formal version is presented as follows. Recall that $\rho_t$ is the PDE solution in \eqref{eq:pde-fixed} and  $\hat \rho_k = m^{-1}\cdot \sum_{i=1}^m \theta_i(k)$ is the empirical distribution of $\theta^{(m)}(k) = \{\theta_i(k)\}_{i=1}^m$. Note that we omit the dependence of $\hat \rho_k$ on $m$ and $\epsilon$ for notational simplicity.
\begin{proposition}[Formal Version of Proposition \ref{prop:discretization0}]
	\label{prop:discretization}
	Let $f: \RR^D\rightarrow \RR$ be any continuous function such that $\norm{f}_\infty \le 1$ and $\lip(f) \le 1$. Under Assumptions \ref{asp:data} and \ref{asp:activation}, it holds that
	\begin{align*}
	&\sup_{\substack{ k\le T/\epsilon \\ (k\in\NN)}}\biggl| \int f(\theta) \,\rd \rho_{k\epsilon}(\theta) - \int f(\theta) \,\rd \hat \rho_{k}(\theta) \biggr| \nonumber\\
	&\quad \le B\cdot e^{BT} \cdot \Bigl( \sqrt{\log(m/\delta)/m}  + \sqrt{\epsilon\cdot \bigl(D + \log(m/\delta)\bigr)}\Bigr)
	\end{align*}
	with probability at least $1-\delta$. Here $B$ is a constant that depends on $\alpha$, $\eta$, $\gamma$, $B_r$, and $B_j \ (j\in \{0, 1, 2\})$.
\end{proposition}
The proof of Proposition \ref{prop:discretization} is based on \cite{mei2018mean, mei2019mean,araujo2019mean}, which  utilizes the propagation of chaos \citep{sznitman1991topics}. Recall that $g(\cdot; \rho)$ is a vector field defined as follows,
\begin{align*}
g(\theta; \rho) = -\alpha \cdot \EE_{\tilde \cD} \Bigl[ \bigl(Q(x; \rho) - r - \gamma \cdot Q(x'; \rho) \bigr) \cdot \nabla_\theta \sigma(x; \theta)\Bigr].
\end{align*}
Correspondingly, we define the finite-width and stochastic counterparts of $g(\theta; \rho)$ as follows,
\begin{align}
\label{eq:g-hat}
\hat g(\theta; \theta^{(m)}) &= -\alpha \cdot \EE_{\tilde \cD} \Bigl[ \bigl(\hat Q(x; \theta^{(m)}) - r -\gamma \cdot \hat Q(x'; \theta^{(m)}) \bigr) \cdot \nabla_\theta \sigma(x; \theta)\Bigr], \\
\label{eq:G}
\hat G_k(\theta; \theta^{(m)}) &= -\alpha\cdot \bigl( \hat Q(x_k; \theta^{(m)}) - r_k -  \gamma\cdot \hat Q(x'_k; \theta^{(m)})\bigr) \cdot\nabla_\theta \sigma(x_k; \theta),
\end{align}
where $(x_k, r_k, x_k') \sim \tilde \cD$.
Following from \cite{mei2019mean,araujo2019mean}, we consider the following four dynamics.
\begin{itemize}[align=left, leftmargin=*]
	\item {\bf Temporal-difference (TD).} We consider the following TD dynamics $\theta^{(m)}(k)$, where $k\in \NN$, with $\theta_i(0) \overset{\iid}{\sim} \rho_0 \ (i\in [m])$  as its initialization,
	\begin{align}
	\label{eq:td}
	\theta_i(k+1) &= \theta_i(k) - \eta\epsilon \cdot \alpha\cdot \Bigl( \hat Q\bigl(x_k; \theta^{(m)}(k)\bigr) - r_k -  \gamma\cdot \hat Q\bigl(x_k'; \theta^{(m)}(k)\bigr)\Bigr) \cdot\nabla_\theta \sigma\bigl(x_k; \theta_i(k) \bigr) \nonumber \\
	&= \theta_i(k) + \eta\epsilon\cdot \hat G_k\bigl(\theta_i(k); \theta^{(m)}(k)\bigr),
	\end{align}
	where $(x_k, r_k, x'_k) \sim \tilde \cD$. Note that this definition is equivalent to \eqref{eq:sgd}.
	
	\item {\bf Expected temporal-difference (ETD).}  We consider the following expected TD dynamics $\breve \theta^{(m)}(k)$, where $k\in \NN$, with $\breve \theta_i(0) = \theta_i(0) \ (i\in [m])$ as its initialization,
	\begin{align}\label{eq:ptd}
	\breve\theta_i(k+1) &= \breve\theta_i(k) - \eta\epsilon \cdot \alpha\cdot \EE_{\tilde \cD}\biggl[\Bigl( \hat Q\bigl(x; \breve\theta^{(m)}(k)\bigr) - r -  \gamma\cdot \hat Q\bigl(x'; \breve\theta^{(m)}(k)\bigr)\Bigr) \cdot\nabla_\theta \sigma\bigl(x; \breve\theta_i(k) \bigr)\biggr] \nonumber \\
	&= \breve \theta_i(k) + \eta\epsilon\cdot \hat g\bigl(\breve \theta_i(k); \breve\theta^{(m)}(k)\bigr).
	\end{align}
	
	\item {\bf Continuous-time temporal-difference (CTTD).} We consider the following continuous-time TD dynamics $\tilde \theta^{(m)}(t)$, where $t\in \RR_+$, with $\tilde \theta_i(0) =\theta_i(0) \ (i\in [m])$ as its initialization,
	\begin{align}
	\label{eq:cttd}
	\frac{\rd }{\rd t} \tilde \theta_i(t) &= -\eta \cdot \alpha \cdot \EE_{\tilde \cD}\biggl[\Bigl( \hat Q\bigl(x; \tilde \theta^{(m)}(t)\bigr) - r -  \gamma\cdot \hat Q\bigl(x'; \tilde \theta^{(m)}(t)\bigr)\Bigr) \cdot\nabla_\theta \sigma\bigl(x; \tilde \theta_i(t) \bigr)\biggr] \nonumber \\
	& = \eta \cdot \hat g\bigl(\tilde\theta_i(t) ; \tilde \theta^{(m)}(t) \bigr ).
	\end{align}
	
	\item {\bf Ideal particle (IP).} We consider the following ideal particle dynamics $\bar \theta^{(m)}(t)$, where $t\in \RR_+$, with $\bar \theta_i(0) =\theta_i(0) \ (i\in [m])$ as its initialization,
	\begin{align}
	\label{eq:idp}
	\frac{\rd }{\rd t} \bar \theta_i(t) &= -\eta \cdot \alpha \cdot \EE_{\tilde \cD}\Bigl[ \bigl( Q(x; \rho_t ) - r -  \gamma\cdot  Q(x'; \rho_t )\bigr) \cdot\nabla_\theta \sigma\bigl(x; \bar \theta_i(t) \bigr) \Bigr] \nonumber \\
	& = \eta \cdot g\bigl(\bar \theta_i(t) ; \rho_t\bigr),
	\end{align}
	where $\rho_t$ is the PDE solution in \eqref{eq:pde-fixed}.
	
\end{itemize}

We aim to prove that $\hat \rho_{k} = m^{-1} \cdot \sum_{i=1}^{m} \delta_{\theta_i(k)}$ weakly converges to $\rho_{k\epsilon}$. For any continuous function $f: \RR^D \rightarrow \RR$ such that $\norm{f}_{\infty} \le 1$ and $\lip(f) \le 1$,
we use the IP, CTTD, and ETD dynamics as the interpolating dynamics,
\begin{align}
\label{eq:dis}
&\overbrace{\biggl| \int f(\theta) \,\rd \rho_{k\epsilon}(\theta) - \int f(\theta) \,\rd \hat \rho_{k}(\theta)\biggr|}^{\displaystyle{\rm PDE-TD}} \nonumber \\
&\quad \le \biggl|\int f(\theta) \,\rd \rho_{k\epsilon}(\theta) - m^{-1} \cdot \sum_{i=1}^{m} f\bigl(\bar \theta_i(k\epsilon) \bigr)\biggr| + \biggl| m^{-1} \cdot \sum_{i=1}^{m} f\bigl(\bar \theta_i(k\epsilon) \bigr) - m^{-1} \cdot \sum_{i=1}^{m} f\bigl(\tilde \theta_i(k\epsilon) \bigr) \biggr| \nonumber \\
&\qquad + \biggl| m^{-1} \cdot  \sum_{i=1}^{m} f\bigl(\tilde \theta_i(k\epsilon) \bigr) - m^{-1} \cdot  \sum_{i=1}^{m} f\bigl(\breve \theta_i(k) \bigr) \biggr| + \biggl| m^{-1} \cdot  \sum_{i=1}^{m} f\bigl(\breve \theta_i(k) \bigr) - m^{-1} \cdot \sum_{i=1}^{m} f\bigl(\theta_i(k) \bigr) \biggr| \nonumber \\
&\quad \le \underbrace{\biggl|\int f(\theta) \,\rd \rho_{k\epsilon}(\theta) - m^{-1} \cdot \sum_{i=1}^{m} f\bigl(\bar \theta_i(k\epsilon) \bigr)\biggr|}_{\displaystyle{\rm PDE-IP}} + \underbrace{\norm[\big]{\bar \theta^{(m)}(k\epsilon) - \tilde\theta^{(m)}(k\epsilon)}_{(m)}}_{\displaystyle{\rm IP-CTTD}} \nonumber \\
&\qquad + \underbrace{\norm[\big]{\tilde \theta^{(m)}(k\epsilon) - \breve\theta^{(m)}(k)}_{(m)}}_{\displaystyle{\rm CTTD-ETD}} + \underbrace{\norm[\big]{\breve \theta^{(m)}(k) - \theta^{(m)}(k)}_{(m)}}_{\displaystyle{\rm ETD-TD}},
\end{align}
where the last inequality follows from the the fact that $\lip(f) \le 1$. Here the norm $\norm{\cdot}_{(m)}$ of $ \theta^{(m)} = \{\theta_i\}_{i=1}^m $ is defined as follows,
\begin{align}
\label{eq:norm-m}
\norm{\theta^{(m)}}_{(m)} = \sup_{i\in[m]} \norm{\theta_i}.
\end{align}
In what follows, we define $B > 0$ as a constant that depends on $\alpha$, $\eta$, $\gamma$, $B_r$, and $B_j \ (j\in \{0,1,2\})$, whose value varies from line to line. We establish the following lemmas to upper bound the terms on the right-hand side of \eqref{eq:norm-m}.

\begin{lemma}[Upper Bound of PDE -- IP]
	\label{lem:pde-ipd}
	Let $f$ be any continuous function such that $\norm{f}_\infty \le 1$ and $\lip(f) \le 1$.
	Under Assumptions \ref{asp:data} and \ref{asp:activation}, it holds for any $f$ that
	\begin{align*}
	\sup_{t\in [0, T]}\Bigl|\int f(\theta)  \,\rd \rho_{t}(\theta) - m^{-1} \cdot \sum_{i=1}^{m} f\bigl(\bar \theta_i(t) \bigr) \Bigr| &\le B\cdot \sqrt{\log(mT/\delta) / m}
	\end{align*}
	with probability at least $1-\delta$.  
\end{lemma}
\begin{proof}
	See \S\ref{sec:pf-pde-ipd} for a detailed proof.
\end{proof}

\begin{lemma}[Upper Bound of IP -- CTTD]
	\label{lem:ipd-cttd}
	Under Assumptions \ref{asp:data} and \ref{asp:activation}, it holds that
	\begin{align*}
	\sup_{t\in [0, T]}\norm[\big]{\bar \theta^{(m)}(t) - \tilde \theta^{(m)}(t)}_{(m)} \le B\cdot e^{BT} \cdot \sqrt{\log(m/\delta) / m}
	\end{align*}
	with probability at least $1-\delta$.
\end{lemma}
\begin{proof}
	See \S\ref{sec:pf-ipd-cttd} for a detailed proof.
\end{proof}

\begin{lemma}[Upper Bound of CTTD -- ETD]
	\label{lem:cttd-ptd}
	Under Assumptions \ref{asp:data} and \ref{asp:activation}, it holds that
	\begin{align*}
	\sup_{\substack{k \le T/\epsilon \\ (k \in \NN)}} \norm[\big]{\tilde \theta^{(m)}(k\epsilon) - \breve\theta^{(m)}(k)}_{(m)} \le B \cdot e^{BT} \cdot \epsilon .
	\end{align*}
\end{lemma}
\begin{proof}
	See \S\ref{sec:pf-cttd-ptd} for a detailed proof.
\end{proof}

\begin{lemma}[Upper Bound of ETD -- TD]
	\label{lem:ptd-td}
	Under Assumptions \ref{asp:data} and \ref{asp:activation}, it holds that
	\begin{align*}
	\sup_{\substack{k \le T/\epsilon \\ (k \in \NN)}} \norm[\big]{\breve\theta^{(m)}(k) - \theta^{(m)}(k)}_{(m)} \le B\cdot e^{BT}\cdot \sqrt{\epsilon\cdot \bigl(D + \log(m/\delta) \bigr)}
	\end{align*}
	with probability at least $1-\delta$
\end{lemma}
\begin{proof}
	See \S\ref{sec:pf-ptd-td} for a detailed proof.
\end{proof}
We are now ready to present the proof of Proposition \ref{prop:discretization}.
\begin{proof}
	Plugging Lemmas \ref{lem:pde-ipd}-\ref{lem:ptd-td} into \eqref{eq:dis}, we have that
	\begin{align*}
    &\sup_{\substack{k \le T/\epsilon \\ (k \in \NN)}}\biggl| \int f(\theta) \,\rd \rho_{k\epsilon}(\theta) - \int f(\theta) \,\rd\hat \rho_{k}(\theta) \biggr| \nonumber\\
    &\quad \le B\cdot e^{BT} \cdot \Bigl( \sqrt{\log(m/\delta)/m}  + \sqrt{\epsilon\cdot \bigl(D + \log(m/\delta)\bigr)}\Bigr)
	\end{align*}
	with probability at least $1-\delta$. Thus, we complete the proof of Proposition \ref{prop:discretization}.
\end{proof}

\subsection{Proofs of Lemmas \ref{lem:pde-ipd}-\ref{lem:ptd-td}}
In this section, we present the proofs of Lemmas \ref{lem:pde-ipd}-\ref{lem:ptd-td}, which are based on \cite{mei2018mean,mei2019mean,araujo2019mean}. We include the required technical lemmas in \S\ref{sec:tech}. Recall that $B > 0$ is a constant that depends on $\alpha$, $\eta$, $\gamma$, $B_r$, and $B_j \ (j\in \{0,1,2\})$, whose value varies from line to line.
\subsubsection{Proof of Lemma \ref{lem:pde-ipd}} \label{sec:pf-pde-ipd} 
\begin{proof}
For the IP dynamics in \eqref{eq:idp}, it holds that $\bar\theta_i(t) \sim \rho_t \ (i\in [m])$ (Proposition 8.1.8 in \cite{ambrosio2008gradient}). Furthermore, since the randomness of $\bar\theta_i(t)$ comes from $\theta_i(0)$ while $\theta_i(0)\ (i\in [m])$ are independent, we have that $\bar\theta_i(t) \overset\iid\sim \rho_t \ (i\in [m])$.
Thus, we have that 
\begin{align*}
\EE_{\rho_{t}} \Bigl[ m^{-1} \cdot \sum_{i=1}^{m} f\bigl(\bar \theta_i(t) \bigr) \Bigr] = \int f(\theta)  \,\rd \rho_{t}(\theta).
\end{align*} 
Let $\theta^{1, (m)} = \{\theta_1, \ldots, \theta_i^1, \ldots, \theta_m\}$ and $\theta^{2, (m)} = \{\theta_1, \ldots, \theta_i^2, \ldots, \theta_m\}$ be two sets that only differ in the $i$-th element. Then, by the condition of Lemma \ref{lem:pde-ipd} that $\norm{f}_\infty \le 1$, we have that
\begin{align*}
\Bigl| m^{-1}\cdot \sum_{j=1}^{m} f(\theta^1_j ) - m^{-1} \cdot \sum_{j=1}^{m} f(\theta^2_j )\Bigr| = m^{-1} \cdot \bigl| f(\theta^1_i ) - f(\theta^2_i) \bigr| \le 2/m.
\end{align*}
Applying McDiarmid's inequality \citep{wainwright2019high}, we have for a fixed $t\in [0, T]$ that
\begin{align}
\label{eq:pf-pi3}
\PP\biggl(\Bigl| m^{-1} \cdot \sum_{i=1}^{m} f\bigl(\bar \theta_i(t) \bigr) - \int f(\theta)  \,\rd \rho_{t}(\theta) \Bigr| \ge p \biggr) \le \exp( -mp^2 / 4 ).
\end{align}
Moreover, we have for any $s, t \in [0, T]$ that
\begin{align*}
&\biggl| \Bigl| m^{-1} \cdot  \sum_{i=1}^{m} f\bigl(\bar \theta_i(t) \bigr) - \int f(\theta)  \,\rd \rho_{t}(\theta) \Bigr| - \Bigl| m^{-1} \cdot \sum_{i=1}^{m} f\bigl(\bar \theta_i(s) \bigr) - \int f(\theta)  \,\rd \rho_{s}(\theta) \Bigr|   \biggr|\nonumber\\
&\quad \le \Bigl| m^{-1} \cdot  \sum_{i=1}^{m} f\bigl(\bar \theta_i(t) \bigr) - m^{-1} \cdot \sum_{i=1}^{m} f\bigl(\bar \theta_i(s) \bigr)\Bigr| + \Bigl| \int f(\theta)  \,\rd \rho_{t}(\theta) - \int f(\theta)  \,\rd \rho_{s}(\theta) \Bigr| \nonumber \\
&\quad \le \norm[\big]{\bar\theta^{(m)}(t) - \bar\theta^{(m)}(s)}_{(m)} + \cW_1(\rho_t, \rho_s)  \nonumber\\
&\quad \le \norm[\big]{\bar\theta^{(m)}(t) - \bar\theta^{(m)}(s)}_{(m)} + \cW_2(\rho_t, \rho_s), 
\end{align*}
where the second inequality follows from the fact that $\lip(f) \le 1$ and \eqref{eq:w-1-dual}.
Applying \eqref{eq:lip-ip} and \eqref{eq:lip-pde} of Lemma \ref{lem:lip-theta}, we have for any $s, t\in [0, T]$ that
\begin{align*}
\biggl| \Bigl| m^{-1} \cdot  \sum_{i=1}^{m} f\bigl(\bar \theta_i(t) \bigr) - \int f(\theta)  \,\rd \rho_{t}(\theta) \Bigr| - \Bigl| m^{-1} \cdot \sum_{i=1}^{m} f\bigl(\bar \theta_i(s) \bigr) - \int f(\theta)  \,\rd \rho_{s}(\theta) \Bigr|   \biggr| \le B\cdot |t - s|.
\end{align*}
Applying the union bound to \eqref{eq:pf-pi3} for $t \in \iota \cdot \{ 0, 1, \ldots, \lfloor T/\iota \rfloor \} $, we have that
\begin{align*}
\PP\biggl(\sup_{t\in [0,T]}\Bigl| m^{-1} \cdot \sum_{i=1}^{m} f\bigl(\bar \theta_i(t) \bigr) - \int f(\theta)  \,\rd \rho_{t}(\theta) \Bigr| \ge p + B \cdot \iota\biggr) \le (T/\iota + 1)\cdot \exp( -mp^2 / 4 ).
\end{align*}
Setting $\iota = m^{-1/2} $ and $p = B\cdot \sqrt{\log(mT/\delta) / m}$, we have that
\begin{align*}
\sup_{t\in [0, T]}\Bigl| m^{-1} \cdot \sum_{i=1}^{m} f\bigl(\bar \theta_i(t) \bigr) - \int f(\theta)  \,\rd \rho_{t}(\theta) \Bigr| \le B\cdot \sqrt{\log(mT/\delta) / m}
\end{align*}
with probability at least $1-\delta$. Thus, we complete the proof of Lemma \ref{lem:pde-ipd}.
\end{proof}

\subsubsection{Proof of Lemma \ref{lem:ipd-cttd}} \label{sec:pf-ipd-cttd} 
\begin{proof}
Recall that $g$ and $\hat g$ are defined in \eqref{eq:g-rho} and \eqref{eq:g-hat}, respectively, that is,
\begin{align*}
g(\theta; \rho) &= -\alpha \cdot \EE_{\tilde \cD} \Bigl[ \bigl(Q(x; \rho) - r - \gamma \cdot Q(x'; \rho) \bigr) \cdot \nabla_\theta \sigma(x; \theta)\Bigr], \\
\hat g(\theta; \theta^{(m)}) &= -\alpha \cdot \EE_{\tilde \cD} \Bigl[ \bigl(\hat Q(x; \theta^{(m)}) - r -\gamma \cdot \hat Q(x'; \theta^{(m)}) \bigr) \cdot \nabla_\theta \sigma(x; \theta)\Bigr].
\end{align*}
Following from the definition of $\tilde \theta_i(t)$ and $\bar\theta_i(t)$ in \eqref{eq:cttd} and \eqref{eq:idp}, respectively, we have for any $i \in [m]$ and $t\in[0,T]$ that
\begin{align}
\label{eq:pf-ic1}
&\norm[\big]{\bar \theta_i(t) - \tilde \theta_i(t)} \nonumber\\
&\quad \le \int_0^t \norm[\bigg]{ \frac{\rd \tilde\theta_i(s)}{\rd s} - \frac{\rd \bar \theta_i(s)}{\rd s} }\,\rd s \nonumber \\
& \quad = \eta \cdot \int_0^t \norm[\Big]{ \hat g\bigl(\tilde \theta_i(s); \tilde\theta^{(m)}(s)\bigr) - g\bigl(\bar \theta_i(s); \rho_s\bigr)}\,\rd s \nonumber\\
&\quad \le \eta \cdot \int_0^t \norm[\Big]{ \hat g\bigl(\tilde \theta_i(s); \tilde\theta^{(m)}(s)\bigr) - \hat g\bigl(\bar \theta_i(s); \bar \theta^{(m)}(s)\bigr)}\,\rd s + \eta  \cdot \int_0^t \norm[\Big]{ \hat g\bigl(\bar \theta_i(s); \bar \theta^{(m)}(s)\bigr) - g\bigl(\bar \theta_i(s); \rho_s\bigr)}\,\rd s \nonumber\\
&\quad \le B\cdot \int_0^t \norm[\big]{\tilde \theta^{(m)}(s) - \bar \theta^{(m)}(s)}_{(m)}\,\rd s + \eta \cdot \int_0^t \norm[\Big]{ \hat g\bigl(\bar \theta_i(s); \bar \theta^{(m)}(s)\bigr) - g\bigl(\bar \theta_i(s); \rho_s\bigr)}\,\rd s, 
\end{align}
where the last inequality follows from \eqref{eq:lip-gh} of Lemma \ref{lem:g-hat}. 
We now upper bound the second term on the right-hand side of \eqref{eq:pf-ic1}. 
Following from the definition of $\hat Q$, $Q$, and $\hat g$ in \eqref{eq:nn-fin}, \eqref{eq:nn-infty}, and \eqref{eq:g-hat}, respectively, we have for any $s \in [0, T]$ and $i \in [m]$ that
\begin{align}
\label{eq:pf-ic2}
&\norm[\Big]{ \hat g\bigl(\bar \theta_i(s); \bar \theta^{(m)}(s)\bigr) - g\bigl(\bar \theta_i(s); \rho_s\bigr)} = \alpha^2 \cdot \norm[\Big]{m^{-1} \cdot \sum_{j=1}^m Z_i^j(s)},
\end{align}
where 
\begin{align*}
Z_i^j(s) = \EE_{\tilde \cD}\biggl[\Bigl(\sigma\bigl(x; \bar\theta_j(s)\bigr) -\int \sigma(x; \theta) \,\rd \rho_s(\theta) - \gamma \cdot  \sigma\bigl(x'; \bar\theta_j(s)\bigr) + \gamma \cdot \int \sigma(x'; \theta) \,\rd \rho_s(\theta) \Bigr) \cdot \nabla_\theta \sigma\bigl(x; \bar \theta_i(s) \bigr)\biggr].
\end{align*}
Following from Assumptions \ref{asp:data} and \ref{asp:activation}, we have that $\norm{Z_i^j(s)} \le B$. When $i \neq j$, following from the fact that $\bar \theta_i(s) \overset{\iid}\sim \rho_s \ (i\in [m]) $, it holds that $\EE[Z_i^j(s) \given \bar \theta_i(s)] = 0$. Following from Lemma \ref{lem:mc}, we have for fixed $s \in [0,T]$ and $i \in [m]$ that
\begin{align}
\label{eq:pf-ic3}
\PP\biggl( \norm[\Big]{m^{-1} \cdot \sum_{j \neq i} Z_i^j(s)} \ge B\cdot (m^{-1/2} + p) \biggr) &= \EE\Biggl[ \PP\biggl( \norm[\Big]{m^{-1} \cdot \sum_{j \neq i} Z_i^j(s)} \ge B\cdot (m^{-1/2} + p)\bigggiven \bar\theta_i(s) \biggr) \Biggr] \nonumber\\&\le \exp(-mp^2).
\end{align}
By \eqref{eq:w-1-dual}, we have that 
\begin{align*}
\sup_{x\in \cX} \Bigl|\int\sigma(x;\theta) \,\rd \rho_s(\theta) - \int\sigma(x;\theta) \,\rd \rho_t(\theta) \Bigr| \le B \cdot \cW_1(\rho_s, \rho_t) \le B \cdot \cW_2(\rho_s, \rho_t) \le B\cdot |s - t|,
\end{align*}
where the last inequality follows from \eqref{eq:lip-pde} of Lemma \ref{lem:lip-theta}.
Thus, following from Assumptions \ref{asp:data} and \ref{asp:activation}, Lemma \ref{lem:lip-theta}, and the fact that $\lip(f g) \le \norm{f}_\infty \cdot \lip(g) + \norm{g}_\infty\cdot \lip(f)$ for any functions $f$ and $g$, we have for any $s, t\in [0, T]$ that
\begin{align*}
\biggl| \norm[\Big]{m^{-1} \cdot \sum_{j \neq i} Z_i^j(s)}  - \norm[\Big]{m^{-1} \cdot \sum_{j \neq i} Z_i^j(t)}\biggr| \le B\cdot |t-s|.
\end{align*}
Applying the union bound to \eqref{eq:pf-ic3} for $ i \in [m]$ and $t\in \iota \cdot \{ 0, 1, \ldots, \lfloor T/\iota \rfloor \}$, we have that
\begin{align*}
\PP\biggl( \sup_{\substack{i\in[m],\\ s\in [0, T]}} \norm[\Big]{m^{-1} \cdot \sum_{j \neq i} Z_i^j(s)} \ge B\cdot (m^{-1/2} + p) + B\iota \biggr) \le m\cdot (T/\iota + 1)\cdot \exp(-mp^2).
\end{align*}
Setting $\iota = m^{-1/2}$ and $p = B\cdot \sqrt{\log(mT/\delta) / m}$, we have that
\begin{align}
\label{eq:pf-ic4}
\sup_{\substack{i\in[m],\\ s\in [0, T]}} \norm[\Big]{m^{-1} \cdot  \sum_{j \neq i} Z_i^j(s)} \le B\cdot \sqrt{\log(mT/\delta) / m}
\end{align}
with probability at least $1-\delta$. When $i=j$, it holds that $\norm{m^{-1}\cdot Z_i^i(s)} \le B / m$ in \eqref{eq:pf-ic2}, which follows from Assumptions \ref{asp:data} and \ref{asp:activation}. Thus, plugging \eqref{eq:pf-ic4} into \eqref{eq:pf-ic2}, we have that
\begin{align}
\label{eq:pf-ic5}
\sup_{\substack{i\in[m],\\ s\in [0, T]}} \norm[\Big]{ \hat g\bigl(\bar \theta_i(s); \bar \theta^{(m)}(s)\bigr) - g\bigl(\bar \theta_i(s); \rho_s\bigr)} &\le \sup_{\substack{i\in[m],\\ s\in [0, T]}} \alpha^2 \cdot \biggl( \norm[\big]{m^{-1} \cdot Z_i^i(s)} + \norm[\Big]{m^{-1} \cdot \sum_{j \neq i} Z_i^j(s)} \biggr) \nonumber \\
&\le B\cdot \sqrt{\log(mT/\delta) / m}
\end{align}
with probability at least $1-\delta$. 

Conditioning on the event in \eqref{eq:pf-ic5}, we obtain from \eqref{eq:pf-ic1} that
\begin{align*}
\norm[\big]{\tilde \theta^{(m)}(t) - \bar \theta^{(m)}(t)}_{(m)} \le B\cdot \int_0^t \norm[\big]{\tilde \theta^{(m)}(s) - \bar \theta^{(m)}(s)}_{(m)}\,\rd s + BT \cdot \sqrt{\log(mT/\delta) / m} 
\end{align*}
for any $t\in [0, T]$.
Following from Gronwall's Lemma \citep{holte2009discrete}, we have that
\begin{align*}
\norm[\big]{\tilde \theta^{(m)}(t) - \bar \theta^{(m)}(t)}_{(m)} &\le B\cdot e^{Bt} \cdot BT \cdot \sqrt{\log(mT/\delta) / m} \nonumber \\
&\le B\cdot e^{BT} \cdot \sqrt{\log(m/\delta) / m},  \qquad \forall t \in [0,T]
\end{align*}
with probability at least $1-\delta$. 
Here the last inequality holds since we allow the value of $B$ to vary from line to line.
Thus, we complete the proof of Lemma \ref{lem:ipd-cttd}
\end{proof}

\subsubsection{Proof of Lemma \ref{lem:cttd-ptd}} \label{sec:pf-cttd-ptd} 
\begin{proof}
By the definition of $ \hat g $, $\breve\theta_i(t)$, and $\tilde \theta_i(t)$ in \eqref{eq:g-hat}, \eqref{eq:ptd}, and \eqref{eq:cttd}, respectively,
it holds that
\begin{align*}
\norm[\big]{\tilde \theta_i(k\epsilon) - \breve\theta_i(k)} &\le \eta \cdot \int_0^{k\epsilon} \norm[\Big]{ \hat g\bigl(\tilde \theta_i(s) ; \tilde \theta^{(m)}(s) \bigr) - \hat g\bigl(\breve \theta_i(\lfloor s/\epsilon\rfloor) ; \breve \theta^{(m)}(\lfloor s/\epsilon \rfloor) \bigr) }\,\rd s \nonumber\\
&\le \eta \cdot \int_0^{k\epsilon} \norm[\Big]{ \hat g\bigl(\tilde \theta_i(s) ; \tilde \theta^{(m)}(s) \bigr) - \hat g\bigl(\tilde \theta_i(\lfloor s/\epsilon\rfloor \cdot \epsilon) ; \tilde \theta^{(m)}(\lfloor s/\epsilon \rfloor \cdot \epsilon )  \bigr) }\,\rd s  \nonumber\\
&\quad + \eta \cdot \sum_{\ell = 0}^{k-1} \norm[\Big]{ \hat g\bigl(\tilde \theta_i(\ell\epsilon) ; \tilde \theta^{(m)}(\ell\epsilon) \bigr) - \hat g\bigl(\breve \theta_i(\ell) ; \breve \theta^{(m)}(\ell) \bigr) } \nonumber\\
&\le B\cdot k\cdot \epsilon^2 + B \cdot \sum_{\ell = 0}^{k-1} \norm[\big]{\tilde \theta^{(m)}(\ell\epsilon) - \breve\theta^{(m)}(\ell)}_{(m)},
\end{align*}
where the last inequality follows from \eqref{eq:lip-gh} of Lemma \ref{lem:g-hat} and \eqref{eq:lip-cttd} of Lemma \ref{lem:lip-theta}. Following from the definition of $\norm{\cdot}_{(m)}$ in \eqref{eq:norm-m}, it holds for any $k \le T/\epsilon \ (k\in \NN)$ that 
\begin{align*}
\norm[\big]{\tilde \theta^{(m)}(k\epsilon) - \breve\theta^{(m)}(k)}_{(m)} \le 
B\cdot T\cdot \epsilon + B \cdot \sum_{\ell = 0}^{k-1} \norm[\big]{\tilde \theta^{(m)}(\ell\epsilon) - \breve\theta^{(m)}(\ell)}_{(m)}.
\end{align*}
Following from the discrete Gronwall's lemma \citep{holte2009discrete}, we have that
\begin{align*}
\sup_{\substack{k\le T/\epsilon \\ (k\in \NN)}} \norm[\big]{\tilde \theta^{(m)}(k\epsilon) - \breve\theta^{(m)}(k)}_{(m)} \le B^2 \cdot T \cdot \epsilon \cdot e^{BT} \le B \cdot e^{BT} \cdot \epsilon,
\end{align*}
where the last inequality holds since we allow the value of $B$ to vary from line to line.
Thus, we complete the proof of Lemma \ref{lem:cttd-ptd}.
\end{proof}

\subsubsection{Proof of Lemma \ref{lem:ptd-td}} \label{sec:pf-ptd-td} 
\begin{proof}
Let $\cG_k = \sigma(\theta^{(m)}(0), z_0, \ldots, z_{k})$ be the $\sigma$-algebra generated by $\theta^{(m)}(0)$ and $z_\ell = (x_\ell, r_\ell, x'_\ell)\ (\ell\le k)$. Recall that $\hat g$ and $\hat G_k$ are defined in \eqref{eq:g-hat} and \eqref{eq:G}, respectively. We have for any $i \in [m]$ and $k\in \NN_+$ that
\begin{align*}
\EE\Bigl[ \hat G_k\bigl(\theta_i(k); \theta^{(m)}(k)\bigr) \Biggiven \cG_{k-1} \Bigr] = \hat g\bigl(\theta_i(k); \theta^{(m)}(k)\bigr).
\end{align*}
Recall that $\theta^{(m)}(k)$ and $\breve\theta^{(m)}(k)$ are the TD and ETD dynamics defined in \eqref{eq:td} and \eqref{eq:ptd}, respectively.
Thus, we have for any $i \in [m]$ and $k \in \NN_+$ that
\begin{align}
\label{eq:pf-pt2}
\norm[\big]{\breve\theta_i(k) - \theta_i(k)} &= \eta\epsilon\cdot \norm[\Big]{ \sum_{\ell=0}^{k-1} \hat G_\ell\bigl(\theta_i(\ell); \theta^{(m)}(\ell)\bigr) - \sum_{\ell=0}^{k-1} \hat g\bigl(\breve \theta_i(\ell); \breve \theta^{(m)}(\ell)\bigr) } \nonumber\\
&\le \eta\epsilon \cdot \norm[\Big]{\sum_{\ell=0}^{k-1} X_i(\ell)} + \eta\epsilon\cdot \sum_{\ell=0}^{k-1}\norm[\Big]{\hat g\bigl(\breve \theta_i(\ell); \breve \theta^{(m)}(\ell)\bigr) - \hat g\bigl(\theta_i(\ell); \theta^{(m)}(\ell)\bigr)} \nonumber\\
&\le \eta\epsilon \cdot \norm[\big]{A_i(k)} + B \epsilon \cdot \sum_{\ell = 0}^{k-1}\norm[\big]{\breve \theta^{(m)}(\ell) - \theta^{(m)}(\ell)}_{(m)},
\end{align}
where the last inequality follows from \eqref{eq:lip-gh} of Lemma \ref{lem:g-hat}, and $X_i(\ell)$ and $A_i(k)$ are defined as
\begin{align*}
X_i(0) &= 0, \\ X_i(\ell) &= \hat G_\ell\bigl(\theta_i(\ell) ;  \theta^{(m)}(\ell)\bigr) - \EE\Bigl[ \hat G_\ell\bigl(\theta_i(\ell); \theta^{(m)}(\ell)\bigr) \Biggiven \cG_{\ell-1} \Bigr] \quad \forall \ell\ge 1, \\
A_i(k) &= \sum_{\ell=0}^{k-1} X_i(\ell).
\end{align*}
Following from \eqref{eq:bound-G} of Lemma \ref{lem:g-hat}, we have that $\norm{X_i(\ell)}\le B$. Thus, the stochastic process $\{A_i(k)\}_{k\in \NN_+}$ is a martingale with $\|A_i(k) - A_i(k-1) \|\le B$.
Applying Lemma \ref{lem:azuma}, we have that
\begin{align}
\label{eq:pf-pt3}
\PP\Bigl( \max_{\substack{k\le T/\epsilon \\ (k\in \NN_+)}} \norm[\big]{A_i(k)} \ge B \cdot \sqrt{T/\epsilon} \cdot (\sqrt{D} + p)\Bigr) \le \exp(-p^2) .
\end{align}
Applying the union bound to \eqref{eq:pf-pt3} for $i \in [m]$, we have that
\begin{align*}
\PP\Bigl( \max_{\substack{i\in [m], \\k\le T/\epsilon \ (k\in \NN_+)}} \norm[\big]{A_i(k)} \ge B \cdot \sqrt{T/\epsilon} \cdot (\sqrt{D} + p)\Bigr) \le m \cdot \exp(-p^2).
\end{align*}
By setting $p = \sqrt{\log(m/\delta)}$, we have that
\begin{align}
\label{eq:pf-pt5}
\norm[\big]{A_i(k)} \le B \cdot \sqrt{T/\epsilon} \cdot \bigl(\sqrt{D} + \sqrt{\log(m/\delta)} \bigr), \quad \forall i\in [m], k \le T/\epsilon \ (k\in \NN_+)
\end{align}
with probability at least $1-\delta$. By \eqref{eq:pf-pt2} and \eqref{eq:pf-pt5}, we have that
\begin{align*}
&\norm[\big]{\breve\theta^{(m)}(k) - \theta^{(m)}(k)}_{(m)} \nonumber\\ &\quad \le B\cdot \sqrt{T\epsilon}\cdot (\sqrt{D} + \sqrt{\log(m/\delta)}) + B\epsilon \cdot \sum_{\ell = 0}^{k-1}\norm[\big]{\breve \theta^{(m)}(\ell) - \theta^{(m)}(\ell)}_{(m)}, \quad
\forall k\le T/\epsilon \ (k\in \NN)
\end{align*} 
with probability at least $1-\delta$.
Applying the discrete Gronwall's Lemma \citep{holte2009discrete}, we have that
\begin{align*}
\norm[\big]{\breve\theta^{(m)}(k) - \theta^{(m)}(k)}_{(m)} &\le B\cdot e^{BT}\cdot B\cdot \sqrt{T\epsilon}\cdot \bigl(\sqrt{D} + \sqrt{\log(m/\delta)} \bigr) \nonumber\\
&\le B\cdot e^{BT}\cdot \sqrt{\epsilon\cdot \bigl(D + \log(m/\delta) \bigr)}, \quad \forall k\le T/\epsilon \ (k\in \NN)
\end{align*}
with probability at least $1-\delta$. Here the last inequality holds since we allow the value of $B$ to vary from line to line. Thus, we complete the proof of Lemma \ref{lem:ptd-td}.
\end{proof}

\subsection{Proof of Lemma \ref{lem:opt-dis}} \label{sec:pf-opt-dis}

\begin{proof}
	Recall that $\hat Q$ and $Q(\cdot; \rho)$ are defined in \eqref{eq:nn-fin} and \eqref{eq:nn-infty}, respectively.
	For notational simplicity, we denote the optimality gaps for $\theta^{(m)} = \{\theta_i\}_{i=1}^m$ and $\rho \in \sP_2(\RR^D)$ by 
	\begin{align}
	L( \theta^{(m)}) &= \EE_\cD \Bigl[ \bigl( \hat Q(x; \theta^{(m)}) - Q^*(x) \bigr)^2 \Bigr], \label{eq:l} \\
	\bar L(\rho) &=\EE_\cD\Bigl[ \bigl( Q(x; \rho) - Q^*(x)\bigr)^2 \Bigr]. \label{eq:l-bar}
	\end{align}
	Recall that $\theta^{(m)}(k)$, $\bar\theta^{(m)}(k\epsilon)$, and $\rho_t$ are the TD dynamics, the IP dynamics, and the PDE solution defined in \eqref{eq:td}, \eqref{eq:idp}, and \eqref{eq:pde-fixed}, respectively.
	It holds for any $k \in \NN$ that 
	\begin{align}
	\label{eq:pf-od1}
	\Bigl| L\bigl(\theta^{(m)}(k)\bigr) - \bar L(\rho_{k\epsilon}) \Bigr| \le \underbrace{ \Bigl| L\bigl(\theta^{(m)}(k)\bigr) - L\bigl(\bar \theta^{(m)}(k\epsilon)\bigr) \Bigr|}_{\displaystyle{\rm (i)}} + \underbrace{ \Bigl| L\bigl(\bar \theta^{(m)}(k\epsilon)\bigr) - \bar L(\rho_{k\epsilon}) \Bigr|}_{\displaystyle{\rm (ii)}}.
	\end{align}
	In what follows, we upper bound the two terms on the right-hand side of \eqref{eq:pf-od1}.
	
	\vskip4pt
	
	\noindent{\bf Upper bounding term (i) of \eqref{eq:pf-od1}.} Following from the definition of $L$ in \eqref{eq:l}, it holds for any $k \in \NN$ that
	\begin{align}
	&\Bigl| L\bigl(\theta^{(m)}(k)\bigr) - L\bigl(\bar \theta^{(m)}(k\epsilon)\bigr) \Bigr|  \nonumber\\
	&\quad = \Biggl| \EE_\cD\biggl[ \Bigl(\hat Q\bigl(x; \theta^{(m)}(k)  \bigr) + \hat Q\bigl(x;\bar \theta_i(k\epsilon) \bigr) - 2Q^*(x) \Bigr) \cdot \Bigl(\hat Q\bigl(x; \theta^{(m)}(k)  \bigr) - \hat Q\bigl(x;\bar \theta_i(k\epsilon) \bigr) \Bigr) \biggr] \Biggr| . \label{eq:ubi1}
	\end{align}
	Following from \eqref{eq:bound-qh}, \eqref{eq:lip-qh}, and \eqref{eq:bound-q} of Lemma \ref{lem:g-hat}, we have for any $k \in \NN$ that
	\begin{align}
	\label{eq:ubi2}
	&\sup_{x\in \cX}\Bigl|\hat Q\bigl(x; \theta^{(m)}(k)  \bigr) + \hat Q\bigl(x;\bar \theta_i(k\epsilon) \bigr) - 2Q^*(x) \Bigr| \le B, \\
	&\sup_{x\in \cX}\Bigl|\hat Q\bigl(x; \theta^{(m)}(k)  \bigr) - \hat Q\bigl(x;\bar \theta_i(k\epsilon) \bigr) \Bigr| \le B\cdot \norm[\big]{\theta^{(m)}(k) - \bar \theta^{(m)}(k\epsilon)}_{(m)}. \label{eq:ubi3}
	\end{align}
	Thus, we have that
	\begin{align}\label{eq:pf-od2}
	&\Bigl| L\bigl(\theta^{(m)}(k)\bigr) - L\bigl(\bar \theta^{(m)}(k\epsilon)\bigr) \Bigr| \nonumber \\
	&\quad \le B \cdot \norm[\big]{\theta^{(m)}(k) - \bar \theta^{(m)}(k\epsilon)}_{(m)} \nonumber\\
	&\quad \le B\cdot e^{BT} \cdot \Bigl( \sqrt{\log(m/\delta)/m}  + \sqrt{\epsilon\cdot \bigl(D + \log(m/\delta)\bigr)}\Bigr), \quad \forall k \le T/\epsilon \ (k\in \NN)
	\end{align}
	with probability at least $1-\delta$. Here the last inequality follows from Lemmas \ref{lem:ipd-cttd}-\ref{lem:ptd-td}.
	
	\vskip4pt
	\noindent{\bf Upper bounding term (ii) of \eqref{eq:pf-od1}.} Let $t =k\epsilon$. It holds for any $t \in [0,T]$ that
	\begin{align}
	\label{eq:pf-od3}
		\Bigl| L\bigl(\bar \theta^{(m)}(t)\bigr) - \bar L(\rho_{t}) \Bigr| \le \biggl| L\bigl(\bar \theta^{(m)}(t)\bigr) - \EE_{\rho_t}\Bigl[L\bigl(\bar \theta^{(m)}(t)\bigr)\Bigr] \biggr| +  \biggl| \EE_{\rho_t}\Bigl[L\bigl(\bar \theta^{(m)}(t)\bigr)\Bigr] - \bar L(\rho_{t}) \biggr|,
	\end{align}
	where the expectation is with respect to $\bar\theta_i(t) \overset\iid\sim \rho_t \ (i\in [m])$.
	For the second term on the right-hand side of \eqref{eq:pf-od3}, following from the fact that $\EE_{\rho_t}[\hat Q(x;\bar\theta^{(m)}(t))] = Q(x; \rho_t)$ for any $x\in \cX$, we have that
	\begin{align}
	\label{eq:pf-od31}
	\biggl| \EE_{\rho_t}\Bigl[L\bigl(\bar \theta^{(m)}(t)\bigr)\Bigr] - \bar L(\rho_{t}) \biggr| &=\biggl| \int \EE_{\rho_t}\Bigl[ \hat Q\bigl(x;\bar \theta^{(m)}(t) \bigr)^2 - Q(x; \rho_t)^2 \Bigr] \,\rd \cD(x) \biggr| \nonumber \\
	&= \biggl| \int {\rm Var}_{\rho_t}\Bigl[  \hat Q\bigl(x;\bar\theta^{(m)}(t) \bigr) \Bigr] \,\rd \cD(x) \biggr| \nonumber\\
	&\le B / m,
	\end{align}
	where the inequality follows from the fact that $\norm{\sigma} \le B$ in Assumption \ref{asp:activation} and the independence of $\bar\theta_i(t) \ (i\in [m])$. Let $\theta^{1, (m)} = \{\theta_1, \ldots, \theta_i^1,\ldots, \theta_m\}$ and $\theta^{2, (m)} =\{ \theta_1, \ldots, \theta_i^2,\ldots, \theta_m\}$ be two sets that only differ in the $i$-th element. It holds that
	\begin{align*}
	\bigl| L(\theta^{1, (m)}) - L(\theta^{2, (m)}) \bigr| \le B \cdot m^{-1} \cdot \EE_\cD\Bigl[\bigl| \sigma(x; \theta_i^1) - \sigma(x; \theta_i^2) \bigr| \Bigr] \le B/ m,
	\end{align*}
	where the first inequality follows from \eqref{eq:ubi1} and \eqref{eq:ubi2} and the second inequality follows from Assumption \ref{asp:activation}. Applying McDiarmid's inequality \citep{wainwright2019high}, we have for a fixed $t\in [0, T]$ that 
	\begin{align}
	\label{eq:pf-od32}
	\PP\Biggl( \biggl| L\bigl(\bar \theta^{(m)}(t)\bigr) - \EE_{\rho_t}\Bigl[L\bigl(\bar \theta^{(m)}(t)\bigr)\Bigr] \biggr| \ge p \Biggr) \le \exp(-mp^2 / B).
	\end{align}
	It holds for any $s, t\in [0, T]$ that
	\begin{align*}
	&\Biggl|\biggl| L\bigl(\bar \theta^{(m)}(t)\bigr) - \EE_{\rho_t}\Bigl[L\bigl(\bar \theta^{(m)}(t)\bigr)\Bigr] \biggr| - \biggl| L\bigl(\bar \theta^{(m)}(s)\bigr) - \EE_{\rho_t}\Bigl[L\bigl(\bar \theta^{(m)}(s)\bigr)\Bigr] \biggr| \Biggr| \\&\quad \le B \cdot \norm[\big]{\bar \theta^{(m)}(t) - \bar \theta^{(m)}(s)}_{(m)} \le B\cdot |t-s|,
	\end{align*}
	where the first inequality follows from \eqref{eq:ubi1}, \eqref{eq:ubi2}, and \eqref{eq:ubi3} and the second inequality follows from \eqref{eq:lip-ip} of Lemma \ref{lem:lip-theta}.
	Applying the union bound to \eqref{eq:pf-od32} for $t \in \iota \cdot \{0, 1, \ldots, \lfloor T/\iota \rfloor \}$, we have that
	\begin{align*}
	\PP\Biggl( \sup_{t\in [0, T]}\biggl| L\bigl(\bar \theta^{(m)}(t)\bigr) - \EE_{\rho_t}\Bigl[L\bigl(\bar \theta^{(m)}(t)\bigr)\Bigr] \biggr| \ge p + B\iota \Biggr) \le (T/\iota + 1)\cdot \exp(-mp^2 / B),
	\end{align*}
	Setting $\iota = m^{-1/2}$ and $p = B\cdot \sqrt{\log(mT\delta)/m}$, we have that
	\begin{align}
	\label{eq:pf-od33}
	\sup_{t\in [0, T]}\biggl| L\bigl(\bar \theta^{(m)}(t)\bigr) - \EE_{\rho_t}\Bigl[L\bigl(\bar \theta^{(m)}(t)\bigr)\Bigr] \biggr| \le  B\cdot \sqrt{\log(mT\delta)/m}
	\end{align}
	with probability at least $1-\delta$. Plugging \eqref{eq:pf-od31} and \eqref{eq:pf-od33} into  \eqref{eq:pf-od3}, noting that $t = k\epsilon$, we have that
	\begin{align} \label{eq:pf-od30}
	\Bigl| L\bigl(\bar \theta^{(m)}(k\epsilon)\bigr) - \bar L(\rho_{k\epsilon}) \Bigr| \le B\cdot \sqrt{\log(mT\delta)/m}, \quad \forall k \le T/\epsilon \ (k\in \NN)
	\end{align}
	with probability at least $1-\delta$.
	
	Plugging \eqref{eq:pf-od2} and \eqref{eq:pf-od30} into \eqref{eq:pf-od1}, we have that 
	\begin{align*}
	\Bigl| L\bigl(\theta^{(m)}(k)\bigr) - \bar L(\rho_{k\epsilon}) \Bigr| \le B\cdot e^{BT} \cdot \Bigl( \sqrt{\log(m/\delta)/m}  + \sqrt{\epsilon\cdot \bigl(D + \log(m/\delta)\bigr)}\Bigr), \quad \forall k\le T/\epsilon \ (k\in \NN)
	\end{align*}
	with probability at least $1-\delta$. Thus, we complete the proof of Lemma \ref{lem:opt-dis}.
\end{proof}

\subsection{Technical Lemmas for \S\ref{sec:discretization}} \label{sec:tech}
In what follows, we present the technical lemmas used in \S\ref{sec:discretization}. Recall that $\hat Q$, $\hat g$, and $\hat G_k$ are defined in \eqref{eq:nn-fin}, \eqref{eq:g-hat}, and \eqref{eq:G}, respectively. Let $B > 0$ be a constant depending on $\alpha$, $ \eta $, $ \gamma $, $ B_r $, and $ B_j \ (j\in \{0,1,2\}) $, whose value varies from line to line.
\begin{lemma}
	\label{lem:g-hat}
	Under Assumptions \ref{asp:data} and \ref{asp:activation}, it holds for any $\theta^{(m)} = \{\theta_i\}_{i=1}^m$ and $\underline \theta^{(m)} = \{\underline \theta_i\}_{i=1}^m$ that
	\begin{align}
	\sup_{x\in \cX} \bigl| \hat Q(x; \theta^{(m)}) \bigr| &\le B, \label{eq:bound-qh}\\ 
	\sup_{x\in \cX}\bigl| \hat Q(x; \theta^{(m)}) - \hat Q(x; \underline\theta^{(m)}) \bigr| &\le B\cdot \norm{\theta^{(m)} - \underline\theta^{(m)} }_{(m)}, \label{eq:lip-qh}\\
	\norm[\big]{\hat G_k(\theta_i; \theta^{(m)})} &\le B, \label{eq:bound-G} \\ 
	\norm[\big]{ \hat G_k(\theta_i; \theta^{(m)}) - \hat G_k(\underline\theta_i; \underline\theta^{(m)}) } &\le B\cdot \norm{\theta^{(m)} - \underline\theta^{(m)} }_{(m)}, \quad   \forall k\in \NN,\label{eq:lip-G}\\
	\norm[\big]{\hat g(\theta_i; \theta^{(m)})} &\le B, \label{eq:bound-gh} \\ 
	\norm[\big]{ \hat g(\theta_i; \theta^{(m)}) - \hat g (\underline\theta_i; \underline\theta^{(m)}) } &\le B\cdot \norm{\theta^{(m)} - \underline\theta^{(m)} }_{(m)}. \label{eq:lip-gh} 
	\end{align}
	Meanwhile, for any $Q \in \cF$, it holds that
	\begin{align}
	\sup_{x\in \cX} \norm[\big]{Q(x)} \le B. \label{eq:bound-q}
	\end{align}
	For any $\rho \in \sP_2(\RR^D)$, it holds that
	\begin{align}
	\norm[\big]{g(\theta; \rho)} &\le B. \label{eq:bound-g}
	\end{align}
\end{lemma}
\begin{proof}
	For \eqref{eq:bound-qh} and \eqref{eq:lip-qh} of Lemma \ref{lem:g-hat},
	following from Assumptions \ref{asp:data} and \ref{asp:activation} and the definition of $\hat Q$ in \eqref{eq:nn-fin}, we have for any $x\in \cX$, $\theta^{(m)}$, and $\underline \theta^{(m)}$ that
	\begin{align*}
	&\bigl| \hat Q(x; \theta^{(m)}) \bigr| \le \alpha \cdot m^{-1} \sum_{i=1}^{m} \bigl| \sigma(x; \theta_i) \bigr| \le B, \\
	&\bigl| \hat Q(x; \theta^{(m)}) - \hat Q(x; \underline\theta^{(m)}) \bigr| \le \alpha \cdot m^{-1} \sum_{i=1}^{m} \bigl| \sigma(x; \theta_i) - \sigma(x; \underline \theta_i) \bigr| \le B\cdot \norm{\theta^{(m)} - \underline \theta^{(m)}}_{(m)}. 
	\end{align*}
	For \eqref{eq:bound-G} and \eqref{eq:lip-G} of Lemma \ref{lem:g-hat},
	following from the definition of $\hat G_k$ in \eqref{eq:G}, we have for any $\theta^{(m)}$ and $\underline \theta^{(m)}$  that
	\begin{align*}
	&\bigl\| \hat G_k(\theta_i; \theta^{(m)})\bigr\| = \alpha \cdot \bigl|\hat Q(x_k; \theta^{(m)}) - r_k- \gamma \cdot \hat Q(x_k'; \theta^{(m)})\bigr| \cdot \norm[\big]{ \nabla_\theta \sigma(x_k; \theta_i)} \le B, \\
	&\norm[\big]{ \hat G_k(\theta_i; \theta^{(m)}) - \hat G_k(\underline\theta_i; \underline\theta^{(m)}) } \\&\quad \le \alpha \cdot \sup_{\theta^{(m)}} \bigl|\hat Q(x_k; \theta^{(m)}) - r_k- \gamma \cdot \hat Q(x_k'; \theta^{(m)})\bigr| \cdot \norm[\big]{ \nabla_\theta \sigma(x_k; \theta_i) - \nabla_\theta \sigma(x_k; \underline \theta_i)} \nonumber\\
	&\qquad + \alpha \cdot \bigl|\hat Q(x_k; \theta^{(m)}) - \gamma \cdot \hat Q(x_k'; \theta^{(m)}) - \hat Q(x_k; \underline \theta^{(m)}) + \gamma \cdot\hat Q(x_k'; \theta^{(m)})  \bigr| \cdot \sup_{\theta_i \in \RR^D}\norm[\big]{ \nabla_\theta \sigma(x_k; \theta_i)} \\
	&\quad \le B \cdot \norm{\theta^{(m)} - \underline \theta^{(m)}}_{(m)}.
	\end{align*}
	The inequalities in \eqref{eq:bound-gh} and \eqref{eq:lip-gh} of Lemma \ref{lem:g-hat} for $\hat g$ follow from the fact that 
	\begin{align*}
	\hat g(\theta_i; \theta^{(m)}) = \EE_{(x_k, r_k, x_k') \sim \tilde \cD}\bigl[G_k(\theta_i; \theta^{(m)})\bigr].
	\end{align*} 
	The inequalities in \eqref{eq:bound-q} and \eqref{eq:bound-g} follow from the definition of $\cF$ and $g$ in \eqref{eq:func-class} and \eqref{eq:g-rho}, respectively. Thus, we complete the proof of Lemma \ref{lem:g-hat}.
\end{proof}

Recall that $\rho_t$ is the PDE solution in \eqref{eq:pde-fixed} and $\tilde\theta^{(m)}(t)$ and $\bar \theta^{(m)}(t)$ are the CTTD and IP dynamics defined in \eqref{eq:cttd} and \eqref{eq:idp}, respectively.
\begin{lemma}
	\label{lem:lip-theta}
	Under Assumptions \ref{asp:data} and \ref{asp:activation},
	it holds for any $s, t \in [0, T]$ that
	\begin{align}
	\norm[\big]{\bar \theta^{(m)}(t) - \bar \theta^{(m)}(s)}_{(m)} &\le B\cdot |t-s|, \label{eq:lip-ip} \\
	\norm[\big]{\tilde \theta^{(m)}(t) - \tilde \theta^{(m)}(s)}_{(m)} &\le B\cdot |t-s|, \label{eq:lip-cttd}\\
	\cW_2(\rho_t, \rho_s) &\le B \cdot |t-s|. \label{eq:lip-pde}
	\end{align}
\end{lemma}
\begin{proof}
	For \eqref{eq:lip-ip} of Lemma \ref{lem:lip-theta},
	by the definition of $\bar \theta_i(t)$ in \eqref{eq:idp} and \eqref{eq:bound-g} of Lemma \ref{lem:g-hat}, we have for any $s, t \in [0, T]$ and $i \in [m]$ that
	\begin{align*}
	\norm[\big]{\bar \theta_i(t) - \bar \theta_i(s)} = \eta\cdot \int_s^t \norm[\Big]{g\bigl(\bar \theta_i(\tau); \rho_\tau \bigr)} \, \rd \tau \le B\cdot |t-s|.
	\end{align*}
	Similarly, for \eqref{eq:lip-cttd} of Lemma \ref{lem:lip-theta}, by the definition of $\tilde \theta_i(t)$ in \eqref{eq:cttd} and \eqref{eq:bound-gh} of Lemma \ref{lem:g-hat}, we have for any $i \in [m]$ and $s, t\in [0, T]$ that $ \norm{\tilde \theta_i(t) - \tilde \theta_i(s)}\le B\cdot |t-s| $.
	
	For \eqref{eq:lip-pde} of Lemma \ref{lem:lip-theta}, following from the fact that $\bar\theta_i(t) \overset\iid\sim \rho_t\ (i\in [m])$ and the definition of $\cW_2$ in \eqref{eq:w2-1}, it holds for any $s, t\in [0, T]$ that
	\begin{align*}
	\cW_2(\rho_t, \rho_s) \le \EE\Bigl[ \norm[\big]{\bar \theta_i(t) - \bar \theta_i(s)}^2 \Bigr]^{1/2} \le B\cdot |t-s|.
	\end{align*}
	Thus, we complete the proof of Lemma \ref{lem:lip-theta}.
\end{proof}

\begin{lemma}[Lemma 30 in \cite{mei2019mean}]
	\label{lem:mc}
	Let $\{X_i\}_{i=1}^m$ be i.i.d.\ random variables with $\norm{X_i} \le \xi$ and $\EE[X_i] = 0$. Then, it  holds for any $p>0$ that
	\begin{align*}
	\PP\bigg( \norm[\Big]{m^{-1} \cdot \sum_{i=1}^{m} X_i} \ge C\xi\cdot (m^{-1/2} + p)\biggr) \le \exp(-mp^2),
	\end{align*}
	where $C > 0$ is an absolute constant.
\end{lemma}

\begin{lemma}[Lemma 31 in \cite{mei2019mean} and Lemma A.3 in \cite{araujo2019mean}]
	\label{lem:azuma}
	Let $X_k\in \RR^D \ (k\in \NN)$ be a martingale with respect to the filtration $\cG_k\ (k\ge 0)$ with $X_0 = 0$. We assume for $\xi > 0$ and any $\lambda \in \RR^D$ that 
	\begin{align*}
	\EE\Bigl[ \exp\bigl( \inp{\lambda}{X_k-X_{k-1}} \bigr) \Biggiven \cG_{k-1} \Bigr] \le \exp\bigl(\xi^2\cdot \norm{\lambda}^2 / 2 \bigr).
	\end{align*}
	Then, it holds that
	\begin{align*}
	\PP\Bigl(\max_{\substack{k\le n\\(k\in \NN)}} \norm{X_k} \ge C \xi \cdot\sqrt{n}\cdot (\sqrt{D} + p)\Bigr) \le \exp(-p^2),
	\end{align*}
	where $C>0$ is an absolute constant.
\end{lemma}

\section{Auxiliary Lemmas}
We use the definition of absolutely continuous curves in $\sP_2(\RR^D)$ in \cite{ambrosio2008gradient}.
\begin{definition}[Absolutely Continuous Curve] \label{def:ac-curve}
	Let $\beta: [a, b] \rightarrow \sP_2(\RR^D)$ be a curve. Then, we say $\beta$ is an absolutely continuous curve if there exists a square-integrable function $f:[a, b] \rightarrow \RR$ such that
	\begin{align*}
	\cW_2( \beta_s, \beta_t) \le \int_{s}^{t} f(\tau) \,\rd \tau
	\end{align*}
	for any $a\le s< t\le b$.	
\end{definition}

Then, we have the following first variation formula.
\begin{lemma}[First Variation Formula, Theorem 8.4.7 in \cite{ambrosio2008gradient}]
	\label{lem:diff}
	Given $\nu \in \sP_2(\RR^D)$ and an absolutely continuous curve $\mu: [0, T] \rightarrow \sP_2(\RR^D)$, let $\beta: [0,1] \rightarrow \sP_2(\RR^D)$ be the geodesic connecting $\mu_t$ and $\nu$. It holds that 
	\begin{align*}
	\frac{\rd }{\rd t}\frac{ \cW_2(\mu_t, \nu)^2}{2} = -\inp{\mu'_t}{\beta'_0}_{\mu_t},
	\end{align*}
	where $\mu'_t = \partial_t \mu_t$, $\beta'_0 = \partial_t \beta_t\given_{t=0}$, and the inner product is defined in \eqref{eq:w-inner}.
\end{lemma}


\begin{lemma}[Talagrand's Inequality, Corollary 2.1 in \cite{otto2000generalization}]
	\label{lem:talagrand}
	Let $\nu$ be $N(0, \kappa \cdot I_D)$. It holds for any $\mu \in \sP_2(\RR^D)$ that
	\begin{align*}
	\cW_2(\mu, \nu)^2 \le 2 D_{\rm KL}(\mu \,\|\, \nu) / \kappa.
	\end{align*}
\end{lemma}

\begin{lemma}[Eulerian Representation of Geodesics, Proposition 5.38 in \cite{villani2003topics}]
	\label{lem:euler}
	Let $\beta: [0, 1] \rightarrow \sP_2(\RR^D)$ be a geodesic and $v$ be the corresponding vector field such that $\partial_t \beta_t = - \Div(\beta_t \cdot v_t)$. It holds that
	\begin{align*}
	\partial_t(\beta_t \cdot v_t) = - \Div(\beta_t \cdot v_t\otimes v_t).
	\end{align*}
\end{lemma}